\pdfoutput=1
\documentclass[11pt]{article}

\usepackage{acl}
\usepackage{times}
\usepackage{latexsym}
\usepackage{tcolorbox}
\usepackage[T1]{fontenc}
\usepackage[utf8]{inputenc}

\usepackage{microtype}

\usepackage{hyperref}
\usepackage{booktabs}
\usepackage{graphicx}
\graphicspath{{./imgs/}}
\usepackage{footmisc}
\usepackage{microtype}
\usepackage{arydshln}
\usepackage{caption}
\usepackage{subcaption}
\usepackage{array, makecell} %
\usepackage{footmisc}
\usepackage{alltt}
\usepackage{floatrow}
\usepackage{pifont}% http://ctan.org/pkg/pifont

\usepackage{tabularx}
\usepackage{adjustbox}

\usepackage{booktabs}
\usepackage[table]{xcolor}
\usepackage{makecell}

\usepackage{multirow}

\usepackage{enumitem}
\usepackage{multirow, colortbl, caption}
\definecolor{lightblue}{RGB}{232, 244, 248}
\definecolor{lightpink}{RGB}{254, 238, 237}

\definecolor{bluelink}{RGB}{0,113,188}
\definecolor{greenlink}{RGB}{0,188,113}
\definecolor{darkblue}{rgb}{0,0,0.55} %

\usepackage{tikz}

\usepackage{collcell}

\usepackage{etoolbox}

\newtoggle{inTableHeader}% Track if still in header of table
\toggletrue{inTableHeader}% Set initial value
\newcommand*{\StartTableHeader}{\global\toggletrue{inTableHeader}}%
\let\OldTabular\tabular%
\let\OldEndTabular\endtabular%
\renewenvironment{tabular}{\StartTableHeader\OldTabular}{\OldEndTabular\StartTableHeader}%

\newcommand*{\MinNumber}{-1.0}%
\newcommand*{\MidNumber}{0.0} %
\newcommand*{\MaxNumber}{1.0}%

\newcommand{\ApplyGradient}[1]{%
  \iftoggle{inTableHeader}{#1}{
    \ifdim #1 pt > \MidNumber pt
        \pgfmathsetmacro{\PercentColor}{max(min(100.0*(#1 - \MidNumber)/(\MaxNumber-\MidNumber),100.0),0.00)} %
        \hspace{-0.33em}\colorbox{yellow!\PercentColor!blue}{#1}
    \else
        \pgfmathsetmacro{\PercentColor}{max(min(100.0*(\MidNumber - #1)/(\MidNumber-\MinNumber),100.0),0.00)} %
        \hspace{-0.33em}\colorbox{blue!\PercentColor!blue}{#1}
    \fi
  }}
\newcolumntype{R}{>{\collectcell\ApplyGradient}c<{\endcollectcell}}

\usepackage{amsmath}
\usepackage{amsfonts,bm}
\usepackage{xspace}

\newcommand{\Ni}{({\em i})~}
\newcommand{\Nii}{({\em ii})~}
\newcommand{\Niii}{({\em iii})~}
\newcommand{\Niv}{({\em iv})~}

\definecolor{mypink3}{cmyk}{0, 0.7808, 0.4429, 0.1412}

\makeatletter   
\newcommand{\sveryshortarrow}[1][3pt]{\mathrel{%
    \vcenter{\hbox{\rule[-.5\fontdimen8\scriptfont3]
               {\scriptratio\dimexpr#1\relax}{\fontdimen8\scriptfont3}}}%
   \mkern-4mu\hbox{\let\f@size\sf@size\usefont{U}{lasy}{m}{n}\symbol{41}}}}
\makeatother

\def\eqref#1{equation~\ref{#1}}
\def\1{\bm{1}}

\def\m1{{\bm{1}}}

\DeclareMathAlphabet{\mathsfit}{\encodingdefault}{\sfdefault}{m}{sl}
\SetMathAlphabet{\mathsfit}{bold}{\encodingdefault}{\sfdefault}{bx}{n}

\usepackage[nameinlink]{cleveref}
\crefformat{section}{\S#2#1#3} % see manual of cleveref, section 8.2.1
\crefname{algorithm}{Alg.}{Algs.}
\crefname{table}{Table}{Tables}
\crefformat{subsection}{\S#2#1#3}
\Crefname{equation}{Eq.}{Eqs.}
\Crefname{figure}{Figure}{Figures}

\usepackage[colorinlistoftodos,prependcaption,textsize=tiny]{todonotes}

\definecolor{darkgreen}{rgb}{0,0.5,0}  
\usepackage{soul}

\usepackage{float}

\usepackage{multirow}% http://ctan.org/pkg/multirow
\usepackage{hhline}% http://ctan.org/pkg/hhline

\usepackage{pgfplots}
\pgfplotsset{compat=1.18}
\usepackage{tikz}
\usetikzlibrary{patterns,positioning}

\definecolor{chartqapro1}{RGB}{30,160,220} % Adjust based on the gradient color (blue)
\definecolor{chartqapro2}{RGB}{50,200,100} % Adjust based on the gradient color (green)

\title{
\textbf{Beyond Static Charts: Can Language and Vision–Language Models Generate Interactive Data Visualization Interfaces?}
}

\author{
\textbf{Mizanur Rahman}\textsuperscript{\textdaggerdbl}
\thanks{Corresponding authors: \{mizanurr,enamulh\}@yorku.ca},
\textbf{Aaryaman Kartha,}\textsuperscript{\textdaggerdbl}, 
\textbf{Enamul Hoque Prince}\textsuperscript{\textdaggerdbl}\footnotemark[1]
\\[2pt]
\textsuperscript{\textdaggerdbl}York University \\
}

\begin{document}
\maketitle

\begin{abstract} 
%Data visualization is central to analytical reasoning and making informed decisions. Real-world analytical workflows relies not only on static charts but more increasingly on interactive visualization interfaces that support filtering, selection, and iterative refinement. 
Data visualization is central to analytical reasoning, but real-world analysis increasingly requires language-driven interactive interfaces rather than static charts.  Although recent large language and vision–language models (LLMs/VLMs) have shown promise in generating static charts from natural language, their ability to generate interactive data visualization interfaces remains largely unexplored due to the lack of benchmarks. We introduce VIS-GEN, a benchmark for evaluating how well LLMs/VLMs can generate interactive visualization interfaces from natural language queries. VIS-GEN comprises 3,042 samples covering diverse analytical intents, including data filtering, temporal analysis, and visualization editing, each paired with dataset metadata and natural language queries that are designed to reflect realistic, goal-driven data exploration scenarios. We benchmark 14 state-of-the-art open-source and closed-source LLMs/VLMs, revealing large performance gaps and frequent failures on queries involving implicit intent, multiple interaction alternatives, and complex editing operations, highlighting interactive interface generation as a key open challenge beyond static chart synthesis. To address this,  we propose a structured multi-stage interface generation framework that decomposes the task into visualization design representation, generation of multiple interface candidates, constraint-aware critique, and self-refinement.  This approach improves the best model’s pass rate by 15.9 percentage points, demonstrating a practical path toward more reliable language-driven interactive visualization systems. We release \textsc{VIS-GEN} at \url{https://github.com/vis-nlp/VIS-GEN}.

\end{abstract}

\section{Introduction}

%  \begin{figure}[t!]
%     \includegraphics[width=\textwidth]{emnlp2020-templates/imgs/text2visfig1_gen1.pdf}
%     \caption{\textbf{Example from the VIS-GEN benchmark.} 
%     \textcolor{darkblue}{\textbf{Input:}} A data table and a natural-language query. 
%     \textcolor{darkgreen}{\textbf{Output:}} An automatically generated interactive visualization interface, including coordinated interaction widgets.\enamul{we need to talk about the fig}}
%     \label{fig:benchmark-sample}
%     \vspace{-2mm}
%     % \vskip -2ex
% \end{figure}

\begin{figure*}[t]
    \centering
    \includegraphics[width=.98\textwidth]{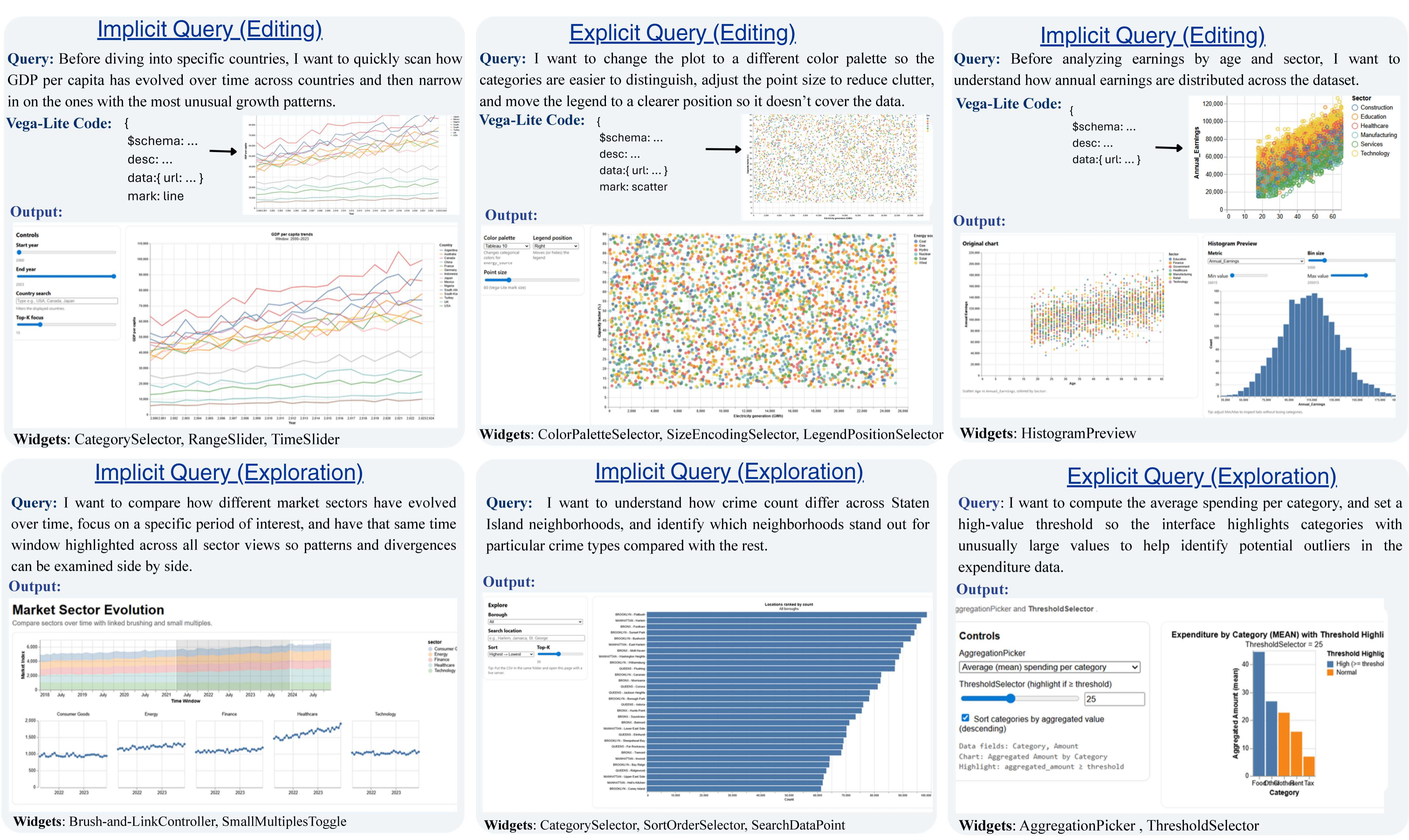}
        \vspace{-4mm}
    \caption{Examples of different query types and expected outputs in VIS-GEN. In both exploration and editing, the model is given a data table and a user query and generates an interactive visualization; in editing, the model is additionally provided with an initial visualization (in Vega-Lite form). Implicit queries describe the analytic goal without specifying operations or controls, whereas explicit queries state the intended operations in plain language.
    %\enamul{in which cases, visualizations are given vs not? Also, do we have any queries referring to visual attributes of the chart (e.g., I would like to to change the red-blue colors, can I change the size of the circles in scatterplot.}
    \vspace{-5mm}
    }
    \label{fig:qa_types}
    
    %\vskip -.5ex
\end{figure*}

Data visualization transforms raw data into visual representations that enable people to make data-driven decisions across domains such as finance, healthcare, and business intelligence~\cite{hoque2022chartquestionansweringstate, rahman2025text2vis}. As a core component of the data science workflow, visualizations are routinely used for exploratory data analysis, comparison, trend identification, and communication of findings~\cite{rahman2025llm}. In practice, real-world analysis rarely relies on static charts alone. Instead, users heavily depend on interactive data visualization interfaces that allow them to select and filter subsets of data, adjust temporal ranges, and iteratively refine visual encodings to support exploration and sensemaking~\cite{yi2007toward}.

Despite their importance, building interactive visualization interfaces remains time-consuming and technically demanding, even for experts~\cite{bach2023challenges}.
%Developing interactive data visualization interfaces is time-consuming and technically demanding, even for experienced practitioners~\cite{bach2023challenges}. 
Unlike static charts, which can often be specified concisely, interactive interfaces require substantial coding to coordinate data transformations, visual encodings, and interaction logic. This process demands expertise in data analysis and visualization frameworks such as D3 or Vega-Lite~\cite{satyanarayan2016vega}, posing a significant barrier for non-technical users~\cite{ali2016big, waskom2021seaborn, bisong2019matplotlib}. While commercial tools such as Tableau and Power BI offer graphical interfaces, effective use still requires design and data transformation skills and provides limited support for flexible, iterative exploration. These limitations raise a fundamental question: \textit{can interactive data visualization interfaces be generated directly from natural language?}

Recent advances in large language and vision–language  models have shown promise in automating visualization-related tasks, particularly static chart generation from natural language queries~\cite{chen2024viseval, rahman2025text2vis}. However, their ability to generate interactive visualization interfaces remains largely unexplored~\cite{vaithilingam2024dynavis}. This task 
%Generating such interfaces 
is substantially more challenging than producing static charts,  as it requires multi-step reasoning over analytical intent, data semantics, visualization design, and interaction constraints. Models must infer appropriate widgets, editing controls, and their coordination with visual encodings—capabilities that existing benchmarks do not capture.

%as it requires supporting multi-step exploration, comparison, and iterative redesign. Beyond interpreting a query and identifying relevant data attributes, systems must infer appropriate interaction widgets and editing controls and coordinate them with visualizations. This process demands joint reasoning over language intent, data semantics, visualization design principles, and interaction constraints, which existing benchmarks fail to capture.

To address these gaps, we introduce VIS-GEN, the first benchmark for evaluating how well vision–language models can generate interactive data visualization interfaces from natural language queries grounded in real datasets. VIS-GEN goes beyond chart generation by requiring models to infer interaction widgets for exploration (e.g., sliders, dropdowns, selectors), editing controls for redesign (e.g., chart type and encoding switches), and their coordination with visual representations.
%(see an example in Figure~\ref{fig:benchmark-sample}). 
The benchmark comprises 3,042 samples spanning diverse analytical intents, including filtering, comparison, temporal analysis, and visualization editing (Figure \ref{fig:qa_types}). Importantly, the queries reflect realistic, goal-driven data exploration scenarios rather than explicit UI instructions, making widget and interface inference a non-trivial reasoning problem.

We benchmark 14 state-of-the-art open-source and closed-source LLMs/VLMs on VIS-GEN and observe substantial performance gaps, particularly for queries involving implicit intent, multiple interaction alternatives, and complex editing operations. Under the direct generation setting, performance remains limited despite near-perfect executability: the strongest closed-source model, Claude-4.5-Opus, achieves only a 41.50\% final pass rate, while the best open-source model reaches 10.91\%.

To move beyond single-pass generation, we introduce a structured multi-stage interface generation framework that decomposes the task into intent extraction, intermediate interface representation, candidate interface synthesis, constraint-aware critique, and validation-driven self-refinement, ultimately producing executable interfaces. 
This approach yields substantial gains, improving final pass rates by 25.1 and 15.9 percentage points for GPT-4o and Claude-4.5-Opus, respectively, and producing more reliable interactive visualizations.

%This approach yields substantial gains in performance, improving GPT-4o’s final pass rate from 11.57\% to 36.69\% and Claude-4.5-Opus from 41.50\% to 58.25\%, and resulting in more accurate, interpretable, and reliable interactive visualizations.

% To support principled generation and analysis, we further propose 
% a structured multi-stage interface generation framework that decomposes the task into %intent parsing, 
% intermediate interface representation, candidate widget and chart generation, self-critique with constraint checking, and candidate ranking prior to producing executable interfaces. Our structured multi-stage framework substantially improves interactive visualization generation, boosting GPT-4o from 11.57\% to 36.69\% in final pass rate and Claude-4.5-Opus from 41.50\% to 58.25\%,
% %increasing interaction control match from 23.40\% to 64.80\%, 
% resulting in significantly more accurate, interpretable, and reliable interactive visualizations. 

In summary, our contributions are threefold:
(1) \textbf{\textsc{VIS-GEN}}, the first benchmark for interactive visualization interface generation from natural language;
%(1)\textbf{ VIS-GEN}, the \textbf{first benchmark} for evaluating language-driven generation of interactive data visualization interfaces grounded in real datasets;
(2) a \textbf{structured interface generation framework} that enables principled reasoning, critique, and selection of interaction widgets and editing controls; and
(3) \textbf{extensive evaluations} of state-of-the-art LLMs/VLMs, uncovering key limitations and highlighting interface-level reasoning as a central challenge for future research in language-driven data visualization.

\vspace{-2mm}
\section{Related Work}
\vspace{-1mm}
\textbf{Text-to-Visualization Benchmarks.} Many existing text-to-visualization benchmarks frame the tasks as natural language–to–SQL translation or direct mappings from text to visualization specifications~\cite{zhong2017seq2sql, luo2021nvbench, srinivasan2021collecting, liu2021advisor, chen2024viseval}. Benchmarks such as WikiSQL~\cite{zhong2017seq2sql} and nvBench~\cite{luo2021nvbench} focus primarily on NL2SQL, implicitly assuming that visualization can be trivially derived from query results, while datasets like NLV-Utterance~\cite{srinivasan2021collecting} and ADVISor~\cite{liu2021advisor} map language to visualization specifications but restrict queries to explicit chart-type mentions and simple analytical intents. VisEval~\cite{chen2024viseval} moves toward evaluating visualization generation but remains limited by a small number of data tables. Finally, Text2Vis~\cite{rahman2025text2vis} improves chart coverage and reasoning complexity, yet still focuses exclusively on static chart generation. 
%As summarized in Table~\ref{tab:datasets_comparison},
None of the existing benchmarks evaluate the generation of interactive visualization interfaces, which motivates our work.

\noindent \textbf{LLMs for Automated Visualization.}
Hybrid methods such as RGVisNet~\cite{song2022rgvisnet} and ADVISor~\cite{liu2021advisor} improved robustness by combining structured pipelines with learned components, while recent LLM-based systems demonstrate strong performance in generating executable visualization code directly from natural language~\cite{hoque2024natural, maddigan2023chat2vis}. For example, Chat2VIS~\cite{maddigan2023chat2vis} applies prompt engineering for chart generation, ChartLlama~\cite{han2023chartllama} focuses on chart understanding via instruction tuning, and Text2Vis-RL~\cite{rahman2025text2vis} introduces a multi-objective reinforcement learning that jointly optimizes textual accuracy, code validity, and visualization quality 
\cite{rahman-etal-2026-aligning}.
%introduces cross-modal agentic inference to jointly refine answers and visualization code. However, all of these approaches primarily target static visualizations. Generating interactive visualization interfaces introduces qualitatively different challenges, including inferring interaction widgets, coordinating state across multiple controls, and supporting iterative editing—capabilities that are not captured by existing benchmarks or modeling formulations.

\noindent\textbf{Generative UI.} Recent advances in VLMs have spurred research on generative user interfaces, where models generate interface structures rather than content alone~\cite{leviathangenerative}. Most prior work focuses on general-purpose UI generation, particularly website and application interfaces, using inputs such as natural language prompts~\cite{chen2025generative, leviathangenerative}, screenshots~\cite{laurenccon2024unlocking, si2025design2code}, or sketches~\cite{li2025sketch2code}, as well as hierarchical or task-model–driven UI synthesis~\cite{gui2025uicopilot, cao2025generative}.
However, these approaches do not account for the data semantics, visual encodings, and analytical intent that fundamentally constrain data visualization interfaces. Work that explicitly targets interactive visualization remains sparse, including ambiguity-resolution widgets~\cite{evizeon}, grammar-based autocompletion for visualization queries~\cite{setlur2020sneak}, SQL-to-interactive visualization systems~\cite{chen2022pi2}, and limited LLM-based visualization editing~\cite{vaithilingam2024dynavis, chen2025generative}. No prior work provides a comprehensive benchmark or systematic evaluation of %language-driven 
interactive visualization interface generation—a gap addressed by VIS-GEN (Tab.~\ref{tab:interactive_positioning}).

 \vspace{-1mm}
\section{\textsc{VIS-GEN}}

We develop VIS-GEN through a carefully designed benchmark construction pipeline that integrates realistic data tables, high-quality human- and LLM-authored queries, executable visualization specifications, and rich annotations, enabling systematic evaluation of both visual correctness and interface-level reasoning. 

% We next describe the benchmark construction process and key characteristics.

\subsection{Data Collection \& Preparation}
To reflect realistic data exploration and visualization editing scenarios, we curate datasets that 
%To ensure that VIS-GEN reflects realistic data exploration and visualization editing scenarios, we curate datasets that 
(i) span diverse real-world topics, (ii) naturally support multiple visualization types, and (iii) afford rich interaction and editing behaviors.
%queries through interface widgets. 
We therefore construct the benchmark using data tables collected from multiple complementary public sources:

\Ni \textbf{Our World in Data (OWID)} ({\href{https://ourworldindata.org}{ourworldindata.org}}) covering global health, economic, social, and demographic indicators with interactive temporal and filtering views.
\Nii \textbf{Organisation for Economic Co-operation and Development (OECD)} ({\href{https://www.oecd.org}{oecd.org}}) providing policy-oriented datasets that support comparison, aggregation, and heterogeneous indicators.
\Niii \textbf{Tableau Public} ({\href{https://public.tableau.com}{public.tableau.com}}) offering practitioner dashboards across business, public policy, and social trends.
\Niv A small fraction of \textbf{synthetic tables} are generated using LLMs to augment structural and semantic diversity, covering rare but important cases underrepresented in real-world sources, such as atypical schemas and complex widget–encoding interactions.

In total, we collect 1,307 data tables across these sources. We then perform manual quality control to remove malformed, redundant, or trivial tables and retain datasets that meaningfully support interactive exploration and visualization editing. The final VIS-GEN collection comprises 966 data tables and is dominated by real-world data, with synthetic tables used primarily to augment diversity. In addition to the data tables, we also curate rich visualization metadata for each instance, including visualization titles, associated widget controls, and interface screenshots, which capture real-world interaction patterns and directly inform the construction of our interaction taxonomy (see Figure~\ref{fig:screenshots}).

% These datasets reflect heterogeneous real-world data conditions. 

% The resulting datasets exhibit characteristics common in practice, such as missing values, multi-variable dependencies, and non-linear relationships, enabling robust evaluation under imperfect and heterogeneous data conditions.
%Importantly, the datasets exhibit characteristics commonly encountered in practical analysis settings—including missing values, multi-variable dependencies, and non-linear relationships—enabling robust evaluation under imperfect and heterogeneous data conditions.
% \enamul{Appendix todos: add some sample screenshots form OECD, OWID and Tableau}
\vspace{-4mm}

\subsection{{Query Generation and Annotations}}  

VIS-GEN supports two complementary interaction scenarios—\emph{exploration} and \emph{editing}—and covers over 40 interaction widgets to capture realistic analytical behavior. We first curate a seed set of 400 expert-authored queries, evenly split between exploration and editing.  Exploration queries require the model to infer both the visualization and interaction strategy directly from data, while editing queries specify analytical refinements to an existing static Vega-Lite visualization.

We then expand this seed set using a multi-step, LLM-assisted refinement process involving GPT-4o, GPT-5.2, Gemini 3.5 Pro, and Claude Sonnet 4.5, producing an additional 2,700 candidates (Tab.~\ref{tab:model-distribution}).
To maintain quality, candidates are generated using few-shot examples drawn from the human-annotated seed set and refined through a multi-stage process before review. Leveraging multiple models increases linguistic and structural diversity. All LLM-generated candidates then undergo manual review by two annotators. Annotators assess each query for realism, clarity, schema grounding, analytical intent, complexity, diversity, and correctness of the associated widget/interface annotations. Details of the human review and revision protocol are provided in Appendix~\ref{app:human_review_protocol}. The initial direct-accept agreement is 82.7\%, with 467 cases flagged for further review. These are manually revised and cross-checked for consistency; candidates are discarded if either annotator rejects them, resulting in 58 removals. In total, 809 queries are either expert-authored or substantially human-revised, forming a significant human-curated subset of the benchmark. The final benchmark contains 3,042 high-quality samples, each paired with a supported widget list, an executable Vega-Lite visualization, and rich metadata, enabling systematic evaluation of both visual generation and interface-level reasoning.

 \begin{figure}[t!]
    \includegraphics[width=\textwidth]{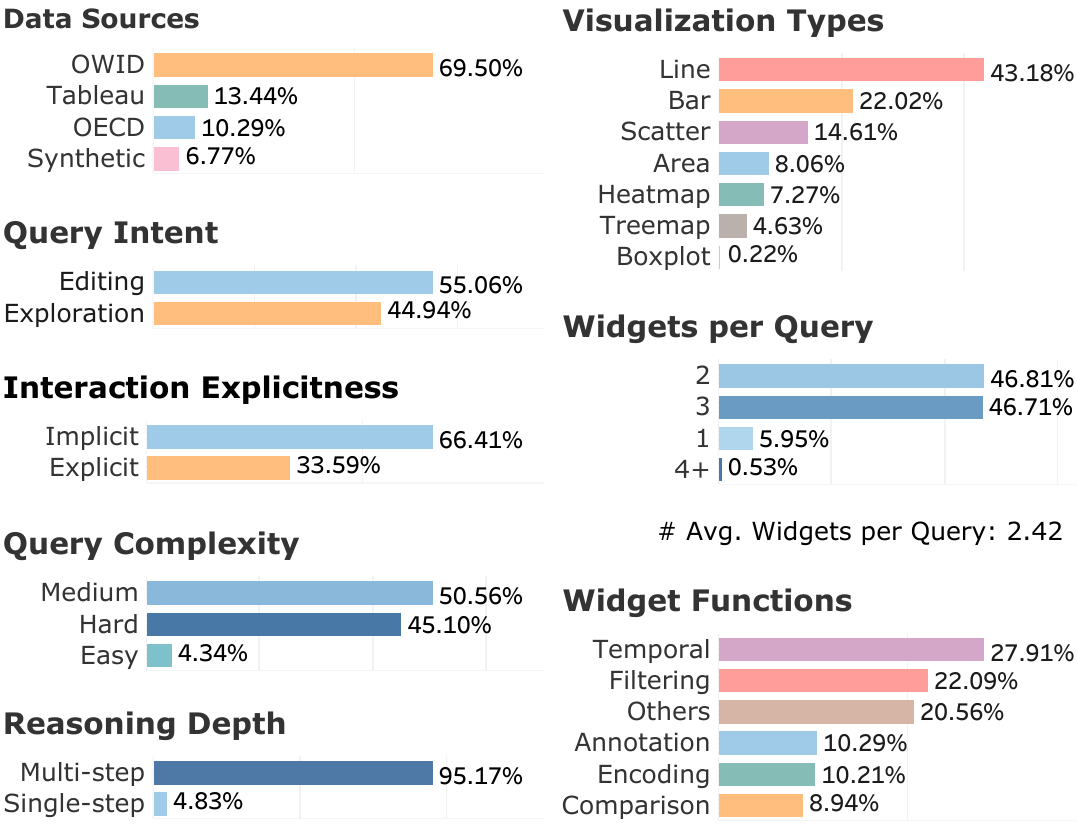}
    \caption{Summary of VIS-GEN dataset query statistics and diversity characteristics.}
    \label{fig:benchmark-statistics}
    \vspace{-2mm}
    % \vskip -2ex
\end{figure}

\subsection{Dataset Analysis}

%\textbf{Data and Topic Diversity.} 
VIS-GEN is constructed primarily from real-world data (93.23\%), with a small fraction of synthetic tables (6.77\%) used to augment rare or underrepresented scenarios, ensuring strong ecological validity alongside controlled diversity (Figure~\ref{fig:benchmark-statistics}). The datasets span diverse topics and schema structures (see~ \Cref{fig:topic-statistics,fig:topic-statistics1}). On average, each table contains 4,133.70 rows (min: 3; max: 184,632) and 9.98 columns (min: 1; max: 167), reflecting both lightweight and large-scale analytical scenarios.

% \enamul{can we add a pie chart of topic distribution in appendix? also any stats of tables? how many rows and columns on average and max.}

\textbf{Analytical Difficulty.} 
VIS-GEN strongly emphasizes non-trivial analytical reasoning. Most queries fall into medium or hard complexity categories, with only a small portion consisting of simple cases. The benchmark is overwhelmingly multi-step (95.17\%), requiring models to reason across multiple analytical operations rather than perform direct lookups or single transformations. In addition, the natural language queries themselves are linguistically rich and complex, with a mean length of 196.46 characters reinforcing the benchmark’s emphasis on realistic, high-cognitive analytical workflows.

\textbf{Interaction Diversity.} 
A defining characteristic of VIS-GEN is its rich and realistic interaction design space. The majority of queries (94.05\%) require multiple coordinated widgets, mirroring real analytical workflows in which users iteratively filter, compare, and refine views. Temporal and filtering interactions are most prevalent, followed by substantial use of comparison, annotation, and encoding-related controls. Importantly, interaction intent is more often implicit (66.41\%) than explicit, forcing models to infer appropriate interface elements from context alone—an essential capability for language-driven visualization systems.

\textbf{Visualization Coverage.}  
VIS-GEN spans a wide range of visualization types. Line ($43.82\%$) and bar charts ($22.02\%$) are most frequent, reflecting common analytical practice, while a meaningful tail of less frequent but analytically important visual forms remains. This distribution ensures that models must generalize across different visualization idioms and reason about how visual encodings interact with data semantics and user intent, rather than overfitting to a narrow set of chart types.

\begin{figure*}[t!]
    \centering
    \includegraphics[width=.98\textwidth]{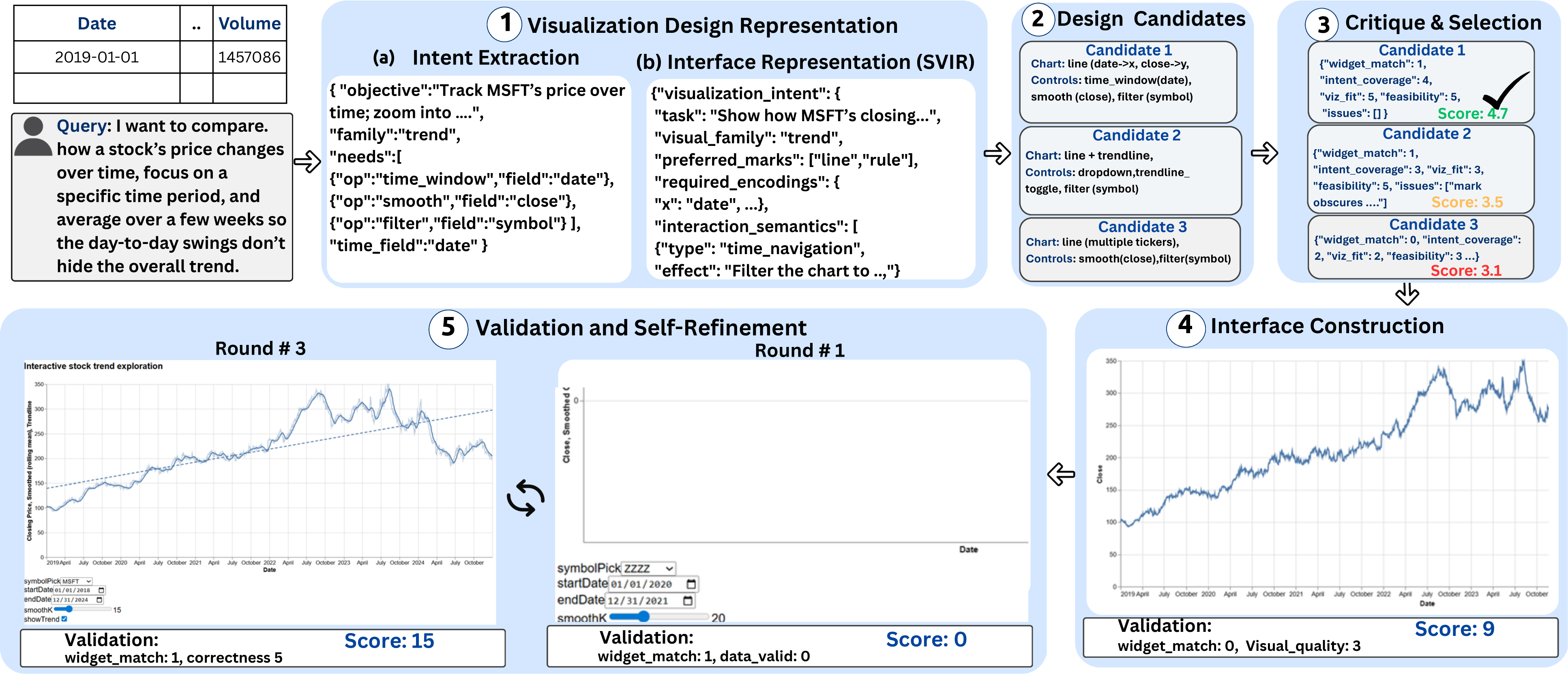}
\caption{Structured multi-stage generation for interactive visualization interfaces:  
Given a dataset and query (and an optional base chart), the model derives a Structured Visualization Interface Representation, generates interface candidates, then selects the best design under constraints, and compiles it into an executable interface. Automated validation detects failures and triggers targeted self-refinement, yielding a functional interactive visualization.
\vspace{-5mm}
}    
%     \caption{
% Multi-stage interface generation framework with candidate design search and automated validation-driven self-refinement.}
    \vspace{-3mm}
    \label{fig:agentic}
    % \vskip -2ex
\end{figure*}

\section{Methodology}
\vspace{-1mm}

\subsection{Task Definition}

We define \textsc{VIS-GEN} as a web-based interactive visualization interface generation task that evaluates a model’s ability to translate natural language analytical queries into executable visualization interfaces grounded in structured data. The dataset consists of $N=3042$ examples, denoted as
$\mathcal{D}=\{(t_i,q_i,s_i,\tau_i,v_i,w_i)\}_{i=1}^{N}$,
where $t_i$ is a data table, $q_i$ is a natural language query,
$s_i$ is an optional base Vega-Lite specification, and
$\tau_i \in \{\text{exploration}, \text{editing}\}$. % The dataset consists of $N{=}3042$ examples, denoted as
% $D = \{(t_i, q_i, s_i, \tau_i, v_i, w_i\}_{i=1}^{N}$,where $t_i$ is a data table, $q_i$ a natural language query, $s_i$ an optional base Vega-Lite specification (editing only), and $\tau_i \in {\text{exploration}, \text{editing}}$. 
Exploration tasks generate interfaces from scratch, whereas editing tasks augment existing visualizations. The model is tasked with generating a Vega-Lite visualization code $v_i$ and a set of interaction widgets $w_i$ from $(t_i, q_i, s_i)$, where the widgets are bound to $v_i$ to control visualization parameters. We adopt Vega-Lite because it provides a widely used declarative grammar with native support for interactive constructs (e.g., parameters and selections), enabling evaluation of language-driven interface reasoning without entangling low-level UI details. Together, $(v_i, w_i)$ define an executable visualization interface.

 \vspace{-1mm}
 
\subsection{Interface Generation Strategies} \label{subsec:evalmethod}

We evaluate two fundamentally different strategies for interactive interface synthesis: direct generation and a structured multi-stage pipeline with validation-driven refinement.

% We evaluate two fundamentally different strategies for interactive interface synthesis: direct generation and structured multi-stage generation with validation-driven refinement.

\noindent\textbf{\Ni Direct Generation.} In the direct setting, the model is prompted to generate the complete interactive interface in a single pass. Given the task inputs, the model jointly produces the visualization specification, interaction widgets, and bindings without any explicit intermediate structure or decomposition. This setting reflects common use of LLMs and serves as a strong baseline for evaluating whether models can internally coordinate analytical intent, visualization design, and interaction behavior.
% This setting reflects the 
% %smost 
% common usage of large language models in practice and serves as a strong baseline for assessing whether models can internally coordinate analytical intent, visualization design, and interaction behavior without external guidance.

% \noindent\textbf{\Nii Structured Multi-Stage Generation.}  While direct generation is appealing for its simplicity, interactive interface design inherently involves multiple interdependent decisions and constraints. To address this, we decompose 
% %propose a structured multi-stage generation framework that decomposes 
% interface synthesis into semantically meaningful stages, inspired by human-centered visualization design methodologies~\cite{munzner2025visualization}, including visualization design representation, design-space exploration, and executable realization (Figure~\ref{fig:agentic}). This structure enables systematic validation and targeted self-refinement, which are difficult to achieve in a single-pass setting. Given a dataset and query, the framework proceeds as follows:

% \noindent \textbf{ (1) Intent Extraction.} The model first infers the underlying analytical intent expressed in the query, including target variables, grouping and comparison dimensions, temporal scope, and required analytical operations. This stage is deliberately agnostic to visualization and UI choices, producing a purely semantic representation of the analytical goal.

\noindent\textbf{\Nii Structured Multi-Stage Generation.}  While direct generation is appealing for its simplicity, interactive interface design inherently involves multiple interdependent decisions and constraints. To address this, we decompose interface synthesis into 
%semantically 
meaningful stages, inspired by human-centered visualization design~\cite{munzner2025visualization}, including design-space exploration, constraint-aware selection, and targeted repair (Figure~\ref{fig:agentic}). Given a dataset and query, the framework proceeds as follows:

%\textbf{(1) Visualization Design Representation.} 
\textbf{(1) Structured Visualization Interface Representation (SVIR)}
The model first infers the underlying analytical intent expressed in the query (target variables, grouping/comparison, temporal scope, and required operations), along with the relevant data fields and interaction needs (\ref{app:svir_schema}). 
%, including target variables, grouping and comparison dimensions, temporal scope, and required analytical operations. 
%This intent specification also identifies the relevant dataset fields and interaction requirements needed to support the analysis.  
The extracted intent is then formalized into a structured JSON representation (SVIR), which specifies required visual encodings (e.g., mark types), permissible interaction types (e.g., filtering, temporal navigation, ranking), and semantic constraints derived from the data schema. %Together, these two representations 
Overall, SVIR defines the complete interface design specification, capturing  \emph{what the interface must support} without committing to specific UI implementations.

 \textbf{(2) Candidate Interface Synthesis.}  The model generates multiple candidate interface designs that satisfy the SVIR defined in Step~(1), each proposing a different combination of visual encodings and abstract interaction controls while respecting dataset-specific field and type constraints. By explicitly generating alternatives, this stage explores the design space rather than committing prematurely to a single solution.
% The model generates multiple candidate interface designs 
% that satisfy the SVIR specification, each representing a different combination of visual encodings and abstract interaction controls. All candidates are required to implement the full set of semantic encodings and interaction intents defined in Step (1), and to respect dataset-specific field and type constraints. By explicitly generating alternatives, this stage explores the design space rather than committing prematurely to a single solution.
% \enamul{do we generate detailed code at this step?}

\textbf{(3) Candidate Critique and Selection.}  
Each candidate interface is evaluated along four complementary dimensions: (i) \textit{Widget Match}, which verifies that all required interaction controls are present; (ii) \textit{Intent Coverage}, which measures how well the design satisfies the user's analytical objectives; (iii) \textit{Visualization Fit}, which assesses the structural alignment between the chart design and the intended analytical task; and (iv) \textit{Feasibility}, which checks schema consistency, field-type compatibility, encoding correctness, and whether the interface can render non-empty data on initial load.  
These criteria are combined using a weighted scoring function over these four dimensions, and the highest-scoring candidate is selected for realization, and candidates violating any hard constraint (missing required interactions, insufficient intent coverage) are discarded (see App. \ref{sec:candidate-selection}). Among the remaining designs, the highest-scoring candidate is selected for realization, introducing explicit and transparent reasoning over design trade-offs that are typically implicit in end-to-end generation.

% Each candidate is evaluated against intent coverage, feasibility, redundancy, and visualization fit. \enamul{why these are useful criteria? how are you combining them? are you doing a weighted approach?} Invalid or low-quality candidates are discarded, and the highest-scoring design is selected for realization. This stage introduces explicit reasoning over design trade-offs that are typically implicit in end-to-end generation.

\textbf{(4) Executable Interface Construction.}   The selected design is translated into a complete interactive HTML interface. Abstract controls are instantiated as concrete UI elements (widgets) and bound to the Vega-Lite visualization through parameters and transforms.

\textbf{ (5) Validation and Self-Refinement.} The generated interface is first subjected to hard validation gates requiring successful execution, presence of required controls, and non-empty data rendering. Each surviving interface is then scored on three criteria, correctness, readability, and visual quality (1-5 scale). The system iteratively refines the interface for up to three refinement rounds or until the scores saturate, and retains the best-scoring result.

% The generated interface is first subjected to hard validation gates that require successful execution, the presence of all required interactive controls, and valid non-empty data rendering. Only interfaces that pass these core checks proceed to quality evaluation. Each surviving interface is then scored on three criteria, correctness, readability, and visual quality, each on a 1--5 scale, producing a maximum possible score of 15. The system iteratively refines the interface until either the maximum number of refinement rounds is reached or the quality score saturates, and the highest-scoring interface is retained as the final output.

% After constructing the executable interface, the system performs a validation-driven self-refinement process. The generated interface is automatically executed and checked for core interface properties, including successful rendering, presence of interactive controls, and absence of critical runtime errors. When violations are detected, the model performs targeted repair while preserving the intended design. This refinement process is repeated for up to two iterations, yielding progressively more reliable interfaces.

\definecolor{human_baseline}{RGB}{255, 245, 170}
\definecolor{open_models_below_4B}{RGB}{185, 235, 255}
\definecolor{open_models_7B_12B}{RGB}{255, 219, 187}
\definecolor{closed_models}{RGB}{240, 240, 240}
\definecolor{chart_specific_models}{RGB}{217, 240, 211}

\begin{table*}[t!]
\centering
\small
\renewcommand{\arraystretch}{1.0}
\setlength{\tabcolsep}{6pt}

\definecolor{closed}{RGB}{230,242,255}
\definecolor{open}{RGB}{255,239,224}

\caption{Automatic evaluation results on VIS-GEN using \textbf{direct generation}. All values are reported in percentage (\%) except Readability, Visual Quality, and Chart Correctness which are on a 5-point scale.
\vspace{-5mm}
}
\vspace{-2mm}
\label{tab:direct}

\scalebox{0.95}{
\begin{tabular}{l|c c c c c c|c}
\toprule
\textbf{Model} &
\makecell{\textbf{Code}\\\textbf{Executability}} &
\makecell{\textbf{Widget}\\\textbf{Match}} &
\makecell{\textbf{Data}\\\textbf{Validity}} &
\makecell{\textbf{Readability}} &
\makecell{\textbf{Visual}\\\textbf{Quality}} &
\makecell{\textbf{Chart}\\\textbf{Correctness}} &
\makecell{\textbf{Final}\\\textbf{Pass}} \\
\midrule

\rowcolor{closed}
\multicolumn{8}{l}{\textbf{Closed-Source Models}} \\

\rowcolor{closed}
GPT-4o & 100 & 24.70 & 46.86 & 2.72 & 2.57 & 2.12 & 12.30 \\

\rowcolor{closed}
GPT-5-mini & 100 & \textbf{61.40} & 36.10 & 3.53 & 3.16 & 2.62 & 28.30 \\

\rowcolor{closed}
O4-mini & 99 & 36.50 & 26.89 & 2.97 & 2.81 & 1.95 & 14.10 \\

\rowcolor{closed}
Gemini-2.5-Pro & 100 & 28.42 & 25.00 & 3.02 & 2.93 & 1.85 & 11.90 \\

\rowcolor{closed}
Claude-4-Sonnet & 99 & 53.40 & 59.10 & 3.62 & 3.48 & 3.05 & 36.20 \\

\rowcolor{closed}
Claude-4.5-Opus & \textbf{100} & 59.30 & \textbf{64.40} & \textbf{3.78} & \textbf{3.68} & \textbf{3.30} & \textbf{41.50} \\

\midrule
\rowcolor{open}
\multicolumn{8}{l}{\textbf{Open-Source Models}} \\

\rowcolor{open}
Qwen-2.5-7B & 100 & 19.60 & 33.52 & 2.16 & 2.07 & 1.70 & 7.20 \\

\rowcolor{open}
Qwen-2.5-14B & 99 & 24.25 & 24.56 & 2.49 & 2.36 & 1.64 & 9.05 \\

\rowcolor{open}
Qwen-2.5-32B & 100 & 25.96 & 31.49 & 2.82 & 2.70 & 1.86 & 10.91 \\

\rowcolor{open}
Qwen2.5-Coder-7B & 100 & 15.59 & 23.50 & 1.99 & 1.87 & 1.54 & 6.90 \\

\rowcolor{open}
Qwen2.5-Coder-14B & 99 & 20.25 & 24.81 & 2.30 & 2.18 & 1.61 & 8.24 \\

\rowcolor{open}
LLaMA-3.1-8B & 99 & 28.81 & 18.01 & 2.60 & 2.27 & 1.67 & 7.79 \\

\rowcolor{open}
CodeLlama-7B & 97 & 20.00 & 16.51 & 2.07 & 1.93 & 1.52 & 5.80 \\

\rowcolor{open}
Mistral-7B & 99 & 22.37 & 18.54 & 2.31 & 1.92 & 1.55 & 6.40 \\

\bottomrule
\end{tabular}
}
\vspace{-2mm}
\end{table*}

\vspace{-1mm}
\section{Evaluation}
\vspace{-1mm}
\subsection{Evaluation Criteria} \label{subsec:evalmetric}
We evaluate generated interactive visualization interfaces along complementary dimensions capturing both technical validity and analytical utility. 
%We evaluate generated interactive visualization interfaces across multiple complementary dimensions capturing both technical validity and analytical utility. 
The evaluation combines automated browser-based testing with VLM-based judgement for scalable and reliable assessment. 
Specifically, we measure: 
\textbf{(i) Executability}, by serving each generated HTML file through a local server and rendering it in a Chromium browser using Playwright~\cite{playwright},
%\footnote{\url{https://playwright.dev}}, 
verifying successful page load, correct Vega-Lite rendering, and absence of fatal runtime errors; 
\textbf{(ii)~Widget presence}, by checking whether the interface contains interactive controls that meaningfully support the analytical intent, accepting reasonable functional equivalents as correct when they satisfy the same intended interaction function.

% by checking whether the interface contains interactive controls that meaningfully support the analytical intent.

% ; and 
% \textbf{(iii) Widget reactivity}, by programmatically interacting with supported widget types (e.g., sliders, dropdowns, buttons, checkboxes, text inputs) and capturing before--after screenshots. For each interaction, GPT-4o judges whether the visualization changes meaningfully in data marks or encodings, ignoring superficial UI changes. Each widget type receives a binary reactivity score, and we define an overall reactivity indicator $R \in \{0,1\}$, where $R=1$ if any tested widget produces a meaningful visualization change.

We further evaluate:
\textbf{(iii) Data validity}, verifying that charts display real plotted data rather than empty or broken visualizations;
\textbf{(iv) Readability}, assessing labeling clarity and usability;
\textbf{(v) Visual quality}, measuring aesthetic coherence and presentation; and
\textbf{(vi) Correctness}, evaluating how well the interface and its interactions satisfy the analytical goal.
Readability, visual quality, and correctness are rated on a 5-point ordinal scale.

An interface is marked \textbf{PASS} 
%if and only 
if executability$=1$,widget\_match$=1$, data\_valid$=1$, and readability, visual quality, and correctness are each $\geq 3$.
All VLM judgments use deterministic decoding (temperature $=0$). GPT-4o serves as the primary judge for all models, with the exception of GPT-4o itself, whose generated outputs are evaluated by Gemini-2.5-Pro to avoid self-evaluation bias.
% GPT-4o serves as the primary judge, with Gemini-2.5-Pro auditing GPT-4o outputs.
%used to audit GPT-4o outputs.

\vspace{-1mm}
\subsection{Models} \label{subsec:data2text}

We evaluate a diverse set of state-of-the-art 
%language 
models to establish strong baselines on VIS-GEN.
Model selection prioritizes code generation ability, structured reasoning, multimodal understanding of tables and natural language, and deployment practicality.
Closed-source models include GPT-4o \cite{openai2024gpt4technicalreport}, GPT-5-mini, o4-mini,
Claude (4-Sonnet, 4.5-Opus)
%Claude-4-Sonnet, Claude-4.5-Opus 
and Gemini 2.5 Pro \cite{geminiteam2024gemini15unlockingmultimodal}. 
For open-source baselines, we focus on models under 15B parameters, including Qwen2.5 (7B,14B,32B Instruct and Coder) and LLaMA-3.1-8B-Instruct \cite{grattafiori2024llama}, CodeLlama-7B \cite{roziere2023code}, and Mistral-7B \cite{jiang2023mistral}.

\begin{table*}[t!]
\centering
\small
\setlength{\tabcolsep}{6pt}
\renewcommand{\arraystretch}{1.0}

\caption{
Comparison of direct generation and the proposed multi-stage framework on VIS-GEN, with ablation results analyzing the contribution of each major component by selectively removing them.
% Comparison between direct generation and the proposed multi-stage framework on VIS-GEN, followed by ablation results of the multi-stage pipeline. We report overall performance and analyze the contribution of each major component by selectively removing them from the system.
\vspace{-4mm}
}
\label{tab:multistage}

\scalebox{0.94}{
\begin{tabular}{l|c c c c c c|c}
\toprule
\textbf{Model}
& \makecell{\textbf{Code}\\\textbf{Executability}}
& \makecell{\textbf{Widget}\\\textbf{Match}}
& \makecell{\textbf{Data}\\\textbf{Validity}}
& \makecell{\textbf{Readability}}
& \makecell{\textbf{Visual}\\\textbf{Quality}}
& \makecell{\textbf{Chart}\\\textbf{Correctness}}
& \makecell{\textbf{Final}\\\textbf{Pass}} \\
\midrule

\rowcolor{gray!12}
\textbf{GPT-4o (Direct)} 
& 100 & 24.70 & 46.86 & 2.72 & 2.57 & 2.12 & 12.30 \\

\rowcolor{green!12}
\textbf{GPT-4o (Multi-Stage)} 
& \textbf{100} & 64.96 & 54.83 & 2.91 & 2.70 & 2.82 & 37.38 \\

\rowcolor{gray!12}
\textbf{Claude-4.5-Opus (Direct)} 
& \textbf{100} & 59.30 & 64.40 & 3.78 & 3.68 & 3.30 & 41.50 \\

\rowcolor{green!12}
\textbf{Claude-4.5-Opus (Multi-Stage)} 
& \textbf{100} & \textbf{67.52} & \textbf{77.15} & \textbf{3.89} & \textbf{3.86} & \textbf{3.66} & \textbf{57.40} \\

\midrule
\rowcolor{orange!12}
\multicolumn{8}{l}{\textbf{Ablation Study (GPT-4o)}} \\

\rowcolor{orange!12}
w/o Intent Extraction        & 100 & 42.48 & 54.50 & 2.78 & 2.62 & 2.55 & 22.78 \\
\rowcolor{orange!12}
w/o SVIR Representation     & 100 & 52.75 & 51.70 & 2.82 & 2.61 & 2.64 & 27.41 \\
\rowcolor{orange!12}
w/o Candidate Critique      & 100 & 54.90 & 51.58 & 2.84 & 2.73 & 2.64 & 26.13 \\
\rowcolor{orange!12}
w/o Self-Refinement         & 99 & 61.35 & 47.56 & 2.73 & 2.58 & 2.49 & 24.92 \\
\rowcolor{orange!12}
Single-Candidate Only       & 100 & 58.10 & 54.00 & 2.86 & 2.63 & 2.66 & 31.61 \\

\rowcolor{orange!12}
w/ Cross-Judge Selection & 100 & 64.20 & 56.90 & 2.95 &2.69  & 2.77  & 36.90\\

\bottomrule
\end{tabular}
}
\vspace{-3mm}
\end{table*}

\begin{table*}[t!]
\centering
\small
\setlength{\tabcolsep}{6pt}
\renewcommand{\arraystretch}{1}

\caption{
Human evaluation on 1000 stratified samples (500 from GPT-4o and 500 from Claude) for both direct and multi-stage generation. Preferred (\%) reflects annotator pairwise preference between the two settings.
\vspace{-5mm}
}
\label{tab:human-eval}

\scalebox{0.94}{
\begin{tabular}{l|c c c c c c|c c}
\toprule
\textbf{Setting}
& \makecell{\textbf{Code}\\\textbf{Executability}}
& \makecell{\textbf{Widget}\\\textbf{Match}}
& \makecell{\textbf{Data}\\\textbf{Validity}}
& \makecell{\textbf{Readability}}
& \makecell{\textbf{Visual}\\\textbf{Quality}}
& \makecell{\textbf{Chart}\\\textbf{Correctness}}
& \makecell{\textbf{Final}\\\textbf{Pass}}
& \makecell{\textbf{Preferred}\\\textbf{(\%)}} \\
\midrule

\rowcolor{gray!12}
\textbf{Direct} 
& 98.5 
& 40.10 
& 56.20 
& 3.26 
& 3.17 
& 2.76 
& 29.90 
& 24.4 \\

\rowcolor{green!12}
\textbf{Multi-Stage} 
& \textbf{99.3} 
& \textbf{71.30} 
& \textbf{69.10} 
& \textbf{3.62} 
& \textbf{3.42} 
& \textbf{3.39} 
& \textbf{50.20} 
& \textbf{75.6} \\

\bottomrule
\end{tabular}
}
\vspace{-3mm}
\end{table*}

\vspace{-1mm}
\subsection{Main Results}
\label{sec:benchmark_results}
\textbf{Direct Generation Results}: Table~\ref{tab:direct} reports results under single-pass interface generation. Among closed-source models, the Claude family leads: \textbf{Claude-4.5-Opus} achieves the highest final pass rate (\textbf{41.50\%}), followed by Claude-4-Sonnet (36.20\%) and GPT-5-mini (28.30\%). These models produce more reliable interfaces across widget matching, data validity, readability, and correctness. In contrast, open-source models perform 
%significantly 
worse. 
The strongest open-source model, Qwen-2.5-32B, reaches only 10.91\% final pass rate, while most others remain below 8\%. This result highlights the severe limitations of open models in generating interactive visualizations.

Across all systems, a consistent pattern emerges: models can usually generate syntactically valid charts but struggle to produce functional interactive interfaces. Despite near-perfect executability, failures often arise from incorrect or incomplete interaction behavior, indicating that the primary challenge lies in coordinating interaction logic with visual encodings, generating valid visualization code.

% Across all systems, a consistent failure pattern emerges: models can usually generate charts, but struggle to make them interactive. Although executability is near perfect, many interfaces fail due to incorrect widget bindings, broken state coordination, or non-functional interactions, and charts sometimes fail to render despite valid HTML. These results show that the core challenge lies not in visualization syntax, but in coordinating widgets, state, and visual encodings.
%into a coherent interactive system.

\vspace{-2mm}
\textbf{Multi-Stage Generation Results.} We evaluate the structured multi-stage framework on the full VIS-GEN benchmark (Table~\ref{tab:multistage}), focusing on two representative extremes: \textbf{GPT-4o}, which performs 
poorly
%near the bottom
under direct generation, and \textbf{Claude-4.5-Opus}, the strongest direct baseline. For GPT-4o, multi-stage generation yields a dramatic improvement, increasing the final pass rate from 12.30\% to 37.38\%, with large gains in widget matching, data validity, and chart correctness and further boosts Opus from 41.50\% to 57.40\%.  
%Applying the same framework to Claude-4.5-Opus produces consistent additional gains, raising performance from  40.84\% to 58.25\%. 
This contrast shows that interface reliability is governed more by the generation strategy than by the underlying model alone. Notably, structured GPT-4o approaches the performance of Opus (direct), while structured Opus establishes the strongest overall results, underscoring the importance of explicit decomposition, constraint-aware critique, and validation-driven refinement for dependable language-driven interactive visualization ( Fig. \ref{fig:vis-comparison}). Although we evaluate the multi-stage framework with GPT-4o and Claude-4.5-Opus, the framework is model-agnostic and can be applied to both closed-source and open-source models. Overall, performance drops most sharply on implicit, hard, multi-step, and multi-widget queries, indicating that failures are driven more by interface-level reasoning (see Appendix \ref{sec:additional-results} for further breakdowns).

% Further breakdowns of model performance by query type, complexity, reasoning depth, and number of widgets are provided in Appendix \ref{sec:additional-results}.%, Tab. \ref{tab:model-dimension-breakdown}.

% Models perform better on explicit queries, single-widget cases, and short reasoning tasks. 

\vspace{-1mm}
\subsection{Ablation Studies}

Table~\ref{tab:multistage} shows that the effectiveness of our multi-stage framework arises from the complementary contributions of its sequential stages, rather than any single step in isolation. The full pipeline achieves a 37.38\% final pass rate, while removing individual stages consistently degrades performance to 22.78--31.61\%, confirming that reliable interface generation benefits from structured decomposition over one-shot synthesis. The largest drops occur when intent extraction or the SVIR representation is removed, underscoring that explicitly modeling analytical goals and required interactions is critical. Removing candidate critique or self-refinement further degrades results. We additionally include a cross-judge variant, where Gemini-2.5-Pro performs Stage 3 selection and refinement evaluation; comparable performance shows that our structured rubric is robust to the choice of self- vs. cross-evaluator.

\vspace{-2mm}
\subsection{Human Evaluation}
\vspace{-1mm}

We conducted a human evaluation with two annotators on 1000 stratified samples (500 from GPT-4o and 500 from Claude~4.5~Opus), covering both direct and multi-stage generation.  Annotators applied the same criteria as the automated evaluation and provided pairwise preferences (Fig. \ref{fig:scoring-rubric}).
As shown in Table~\ref{tab:human-eval}, the multi-stage framework achieves large improvements over direct generation across widget matching, data validity, chart correctness, and final pass rate, and is strongly preferred by annotators (75.6\% of cases). % We also  manually tested widget reactivity by interacting with generated widgets and found that interfaces passing the final criteria almost always exhibit the expected interactions (93/94). 
We also conducted a larger widget-reactivity audit on all PASS samples from 1000 multi-stage generations, since samples that already fail any required criterion remain non-PASS regardless of reactivity. Among 502 PASS samples, 493/502 (98.21\%) exhibited correct widget–visualization linkage, confirming that interfaces satisfying the final criteria almost always support the expected interactions.
Finally, we found strong correlation between human and model judgments for each metric, with high Pearson ($r$) and Spearman ($\rho$) coefficients ranging from 0.85 to 1.00 (see Table~\ref{tab:human-judge-corr}).

\vspace{-1mm}

\vspace{-1mm}
\subsection{Error Analysis}
\label{sec:error_analysis}
% \enamul{shorten this section while keeping main points. refer to example fig for each error}
\label{sec:error_analysis}
\vspace{-1mm} 
We conducted a qualitative error analysis on 250 samples to identify key error patterns (see fig.\ref{fig:error-taxonomy}):

\textbf{Missing or incorrect widgets.} In direct generation, models often produce static charts, include too few controls, or add widgets that do not satisfy the query, causing failures despite plausible visuals.

\textbf{Data validity failures.} Some interfaces produce empty or incorrect views due to hallucinated field names, type mismatches (e.g., numeric vs. categorical), incorrect aggregation or filtering, or malformed parameter definitions, including incompatible assumptions across visualization grammars.

\textbf{Visual and semantic mismatches.} Some outputs produce readable charts but violate visualization design principles, e.g., using pie charts for temporal trends, resulting in low chart correctness.

\textbf{Broken bindings and state updates.} Even when correct widgets are present, interactions may fail to update the visualization state due to disconnected parameters, missing references between parameters and encodings or filters, or incorrect signal wiring, resulting in non-responsive
%or empty 
interfaces.

\begin{table*}[t]
\centering
\caption{
Model performance breakdown by query dimensions under Direct and Multi-stage settings.
Metrics: WM (Widget Match, \%), DV (Data Valid, \%), CC (Chart Correctness, 1--5), PASS (Pass Rate, \%).
}
\resizebox{\textwidth}{!}{%
\begin{tabular}{l|cc|cc|cc|cc}
\toprule
\multirow{2}{*}{\textbf{Category}}
& \multicolumn{2}{c|}{\textbf{Widget Match}}
& \multicolumn{2}{c|}{\textbf{Data Validity}}
& \multicolumn{2}{c|}{\textbf{Chart Correctness}}
& \multicolumn{2}{c}{\textbf{Final Pass Rate}} \\
\cmidrule(lr){2-3} \cmidrule(lr){4-5} \cmidrule(lr){6-7} \cmidrule(lr){8-9}
& \textbf{Direct} & \textbf{Multi}
& \textbf{Direct} & \textbf{Multi}
& \textbf{Direct} & \textbf{Multi}
& \textbf{Direct} & \textbf{Multi} \\
\midrule

% ===========================
% GPT-4o
% ===========================
% \multicolumn{9}{l}{\rowcolor{black!5}\textbf{GPT-4o}} \\

\rowcolor{black!5}
\multicolumn{9}{l}{\textbf{GPT-4o}} \\

\multicolumn{9}{l}{\textbf{\textit{Query intent}}} \\
Exploration & 15.00 & 61.81 & 48.65 & 51.94 & 2.09 & 2.94 & 10.17 & 35.26 \\
Editing     & 32.60 & 67.52 & 45.37 & 57.19 & 2.14 & 2.72 & 14.03 & 39.10 \\

\multicolumn{9}{l}{\textbf{\textit{Query complexity}}} \\
Easy   & 31.82 & 71.21 & 56.06 & 55.30 & 2.40 & 2.92 & 18.18 & 39.39 \\
Medium & 40.05 & 68.40 & 44.99 & 53.97 & 2.20 & 2.83 & 18.99 & 37.71 \\
Hard   & 6.78 & 60.50 & 48.03 &55.76 & 1.99 & 2.80 & 4.23 & 36.81 \\

\multicolumn{9}{l}{\textbf{\textit{Interaction explicitness}}} \\
Explicit & 48.34 & 81.51 & 33.66 & 46.48 & 2.07 & 2.77 & 18.59 & 38.06 \\
Implicit & 12.72 & 56.58 & 53.51 & 59.06 & 2.15 & 2.85 & 9.11 & 37.03 \\

\multicolumn{9}{l}{\textbf{\textit{\# Widgets required}}} \\
Single (1) & 56.35 & 78.45 & 70.17 & 77.90 & 3.10 & 3.15 & 47.51 & 50.83 \\
Multi ($\geq$2) & 22.68 & 64.10 & 45.37 & 53.37 & 2.06 & 2.80 & 10.10 & 36.53 \\

\multicolumn{9}{l}{\textbf{\textit{Reasoning length}}} \\
Short & 55.78 & 73.47 & 61.90 & 64.63 & 2.70 & 3.05 & 34.69 & 42.18 \\
Long  & 23.11 & 64.53 & 46.08 & 54.34 &2.09 & 2.81 & 11.20 & 37.13 \\

\midrule

% ===========================
% Claude-4.5-Opus
% ===========================
% \multicolumn{9}{l}{\rowcolor{black!5}\textbf{Claude-4.5-Opus}} \\

\rowcolor{black!5}
\multicolumn{9}{l}{\textbf{Claude-4.5-Opus}} \\

\multicolumn{9}{l}{\textbf{\textit{Query intent}}} \\
Exploration & 54.94 & 60.00 & 69.70 & 86.10 & 3.48 & 3.91 & 43.70 & 55.96 \\
Editing     & 62.87 & 73.67 & 60.00 & 69.85 & 3.15 & 3.45 & 39.04 & 58.57 \\

\multicolumn{9}{l}{\textbf{\textit{Query complexity}}} \\
Easy   & 75.76 & 77.27 & 77.30& 84.10 & 3.65 & 3.93 & 55.30 & 61.36 \\
Medium & 73.20 & 75.03& 63.00 &73.86 & 3.43 & 3.72 & 50.13 & 59.62 \\
Hard   & 42.13 & 58.16 & 64.70 & 80.17 & 3.12 & 3.58 & 29.73 & 54.52 \\

\multicolumn{9}{l}{\textbf{\textit{Interaction explicitness}}} \\
Explicit & 79.06 & 82.40 & 58.50& 70.74 & 3.40& 3.82 & 50.10 & 63.89\\
Implicit & 49.30 & 60.10 &  67.40& 80.40& 3.26 & 3.59 & 36.78 & 54.11 \\

\multicolumn{9}{l}{\textbf{\textit{\# Widgets required}}} \\
Single (1) & 89.00 & 79.60 & 76.20 & 86.19 & 4.20& 4.28 & 65.75 & 68.51\\
Multi ($\geq$2) & 57.40 & 66.76 & 63.60 & 76.58 & 3.25 & 3.62 & 40.00 & 56.70 \\

\multicolumn{9}{l}{\textbf{\textit{Reasoning length}}} \\
Short & 85.00 & 83.57 & 76.20& 88.44 & 3.82& 4.07 & 58.50 & 66.67 \\
Long  & 57.98 & 66.70 & 63.80 & 76.58 & 3.27 & 3.64 & 40.60& 56.93 \\

\bottomrule
\end{tabular}}
\label{tab:model-dimension-breakdown}
\vspace{-2mm}
\end{table*}

\begin{table*}[t!]
\centering
\renewcommand{\arraystretch}{1.2}
\small
\setlength{\tabcolsep}{3.5pt}
\caption{\textbf{Comparison of VIS-GEN with prior visualization generation systems and benchmarks.}
\ding{51} = Yes, $\triangle$ = Partial, \ding{55} = No.}
\label{tab:interactive_positioning}

\resizebox{0.98\textwidth}{!}{
\begin{tabular}{l c c c c c c c c}
\toprule
\textbf{Method} &
\makecell{\textbf{Input}} &
\makecell{\textbf{Output}} &
\makecell{\textbf{Eval./Benchmark} \\ \textbf{Size}} &
\makecell{\textbf{Supports} \\ \textbf{Widgets}} &
\makecell{\textbf{Multi-Widget} \\ \textbf{Coord.}} &
\makecell{\textbf{Editing} \\ \textbf{Tasks}} &
\makecell{\textbf{Implicit} \\ \textbf{Queries}} &
\makecell{\textbf{Executable} \\ \textbf{UI}} \\
\midrule

\rowcolor[HTML]{F2F2F2}
\textbf{PI2} \cite{chen2022pi2} &
\makecell{SQL / example \\ queries} &
\makecell{Interactive \\ interface} &
\makecell{N/A} &
$\triangle$ &
$\triangle$ &
\ding{55} &
\ding{55} &
\ding{51} \\

\rowcolor[HTML]{F2F2F2}
\textbf{DynaVis} \cite{vaithilingam2024dynavis} &
\makecell{NL edit / \\ widget commands} &
\makecell{Widget-augmented \\ visualization} &
\makecell{24-user \\ study} &
$\triangle$ &
$\triangle$ &
\ding{51} &
\ding{55} &
\ding{51} \\

\rowcolor[HTML]{F7F7F7}
\textbf{VisEval} \cite{chen2024viseval} &
\makecell{NL + data} &
\makecell{Static \\ chart} &
\makecell{2,524 \\ queries} &
\ding{55} &
\ding{55} &
\ding{55} &
$\triangle$ &
\ding{55} \\

\rowcolor[HTML]{F7F7F7}
\textbf{Text2Vis} \cite{rahman2025text2vis} &
\makecell{NL + data} &
\makecell{Static chart \\ + answer} &
\makecell{1,985 \\ samples} &
\ding{55} &
\ding{55} &
\ding{55} &
\ding{51} &
\ding{55} \\

\midrule
\rowcolor[HTML]{E5F1FB}
\textbf{VIS-GEN (Ours)} &
\makecell{NL + data} &
\makecell{Full interactive \\ interface} &
\makecell{3,042 \\ samples} &
\ding{51} &
\ding{51} &
\ding{51} &
\ding{51} &
\ding{51} \\

\bottomrule
\end{tabular}}
\vspace{-2mm}
\end{table*}

\vspace{-2  mm}
\section{Conclusion}
\vspace{-1mm}
We introduce \textbf{VIS-GEN}, the first large-scale benchmark for language-driven generation of interactive data visualization interfaces, moving beyond the static-chart focus of prior work. Our evaluation of open- and closed-source models reveals fundamental limitations in interface-level reasoning, particularly for implicit intent, coordinated interactions, and complex editing tasks. To address these challenges, we propose a structured multi-stage generation framework that decomposes interface synthesis into intent modeling, semantic representation, candidate generation and critique, and validation-driven refinement. This approach substantially improves the performance, enabling models to produce more usable and faithful interactive visualization interfaces. Together, VIS-GEN and our multi-stage framework establish a strong foundation for advancing interface-level reasoning in VLMs.

% , bringing us closer to truly usable language-driven analytical systems.
% \section{Conclusion}

% In this work, we introduced VIS-GEN, the first large-scale benchmark for evaluating language-driven generation of interactive data visualization interfaces, addressing a major limitation of prior work that has focused almost exclusively on static charts. Through extensive evaluation of state-of-the-art open-source and closed-source models, we demonstrate that current systems remain fundamentally limited in interface-level reasoning, particularly for tasks requiring implicit intent understanding, coordinated interactions, and complex editing operations. We further proposed a structured multi-stage interface generation framework that decomposes the problem into intent reasoning, semantic interface representation, candidate synthesis and critique, and validation-driven self-refinement. Empirically, this framework yields substantial improvements in reliability and analytical fidelity, improving the final pass rate from 11.57\% to 36.69\% when applied to GPT-4o and matching or surpassing the strongest closed-source baselines. Ablation studies and human evaluations confirm the importance of each pipeline component and the resulting gains in stability, usability, and analytical correctness. Together, VIS-GEN and our framework establish a strong foundation for future research on interface-level reasoning in language and vision–language models and advance the goal of truly usable, language-driven analytical systems.

\section*{Limitations}
While VIS-GEN provides the first benchmark for evaluating language-driven interactive visualization generation, it has several limitations. First, although the benchmark is constructed from diverse real-world and synthetic data, it represents a curated subset of visualization tasks and does not aim to exhaustively cover all analytical contexts. Second, while VIS-GEN supports a wide range of commonly used visualization types, it does not include some less common or more complex visual encodings (e.g., parallel coordinates or large multi-view dashboards). Finally, VIS-GEN is instantiated using Vega-Lite as the visualization grammar, a widely adopted declarative framework supported across both JavaScript and Python ecosystems (e.g., via Altair). The benchmark does not evaluate generation for other interactive visualization libraries, but emphasizes reasoning over analytical intent and interaction structure rather than low-level UI implementation. Similar evaluation principles could be extended to other visualization libraries with comparable interaction primitives (e.g., D3.js).

\section*{Ethical Considerations}

All datasets used in this work are obtained from publicly available sources, including Our World in Data, OECD, and Tableau Public, or generated synthetically.  The study focuses on evaluating the technical capabilities of vision–language models for interactive visualization generation and does not involve human subjects or user data. As such, we do not anticipate direct ethical risks associated with data collection or use. Nevertheless, like other generative systems, models evaluated in this work could potentially be misused to produce misleading or incorrect visualizations if deployed without appropriate human oversight. We encourage responsible deployment practices and careful validation in real-world applications. Finally, we used AI-based writing assistants only to improve the presentation of the paper.

\section*{Acknowledgements}
This research was supported by the Natural Sciences and Engineering Research Council (NSERC),
Canada, Canada Foundation for Innovation, Compute Canada, and the CIRC grant on Inclusive and
Accessible Data Visualizations and Analytics.

\bibliography{text2vis}
\newpage
% \cleardoublepage
\appendix
\section{Appendices}
\label{app:Appendice}

\subsection{Dataset Collection}

Table \ref{tab:data-sources} shows the distribution of data tables from different sources in VIS-GEN.

\begin{table}[t!]
\centering
\small
\renewcommand{\arraystretch}{1.3} 

\caption{Distribution of data tables from different sources in VIS-GEN.}
\label{tab:data-sources}

% Forces the table to the exact width of the column
\begin{tabular*}{\columnwidth}{l @{\extracolsep{\fill}} r}
\toprule
\textbf{Source} & \textbf{Count (\%)} \\
\midrule
Our World in Data (OWID) & 710 (73.50\%) \\
\rowcolor{gray!10}
Tableau Public           & 107 (11.08\%) \\
OECD                     & 94 (9.73\%) \\
\rowcolor{gray!10}
Synthetic                & 55 (5.69\%) \\
\midrule
\textbf{Total}           & \textbf{966 (100.00\%)} \\
\bottomrule
\end{tabular*}
\end{table}

 \begin{figure}[t!]
    \includegraphics[width=\textwidth]{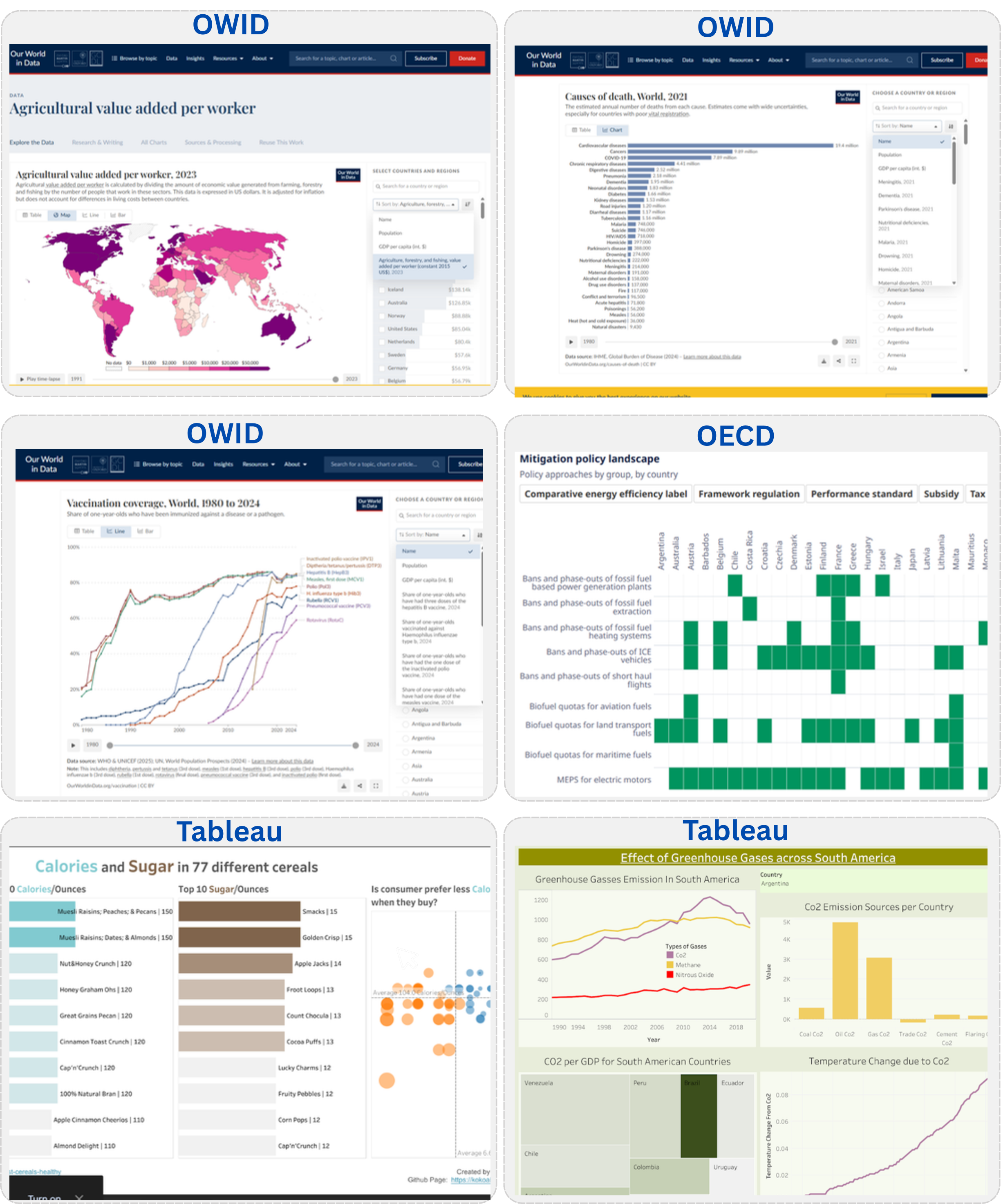}
    \caption{
We collect real interactive visualizations from \textsc{Our World in Data}, \textsc{OECD}, and \textsc{Tableau Public}, extracting both the underlying data tables and rendered interface screenshots. From these, we identify the visual encodings and concrete interaction widgets used in practice, and construct natural-language queries that encode both analytical goals and required interactions, grounding \textsc{VIS-GEN} in realistic visualization workflows.
}
\label{fig:screenshots}
    \vspace{-2mm}
    % \vskip -2ex
\end{figure}

 \begin{figure}[t!]
    \includegraphics[width=\textwidth]{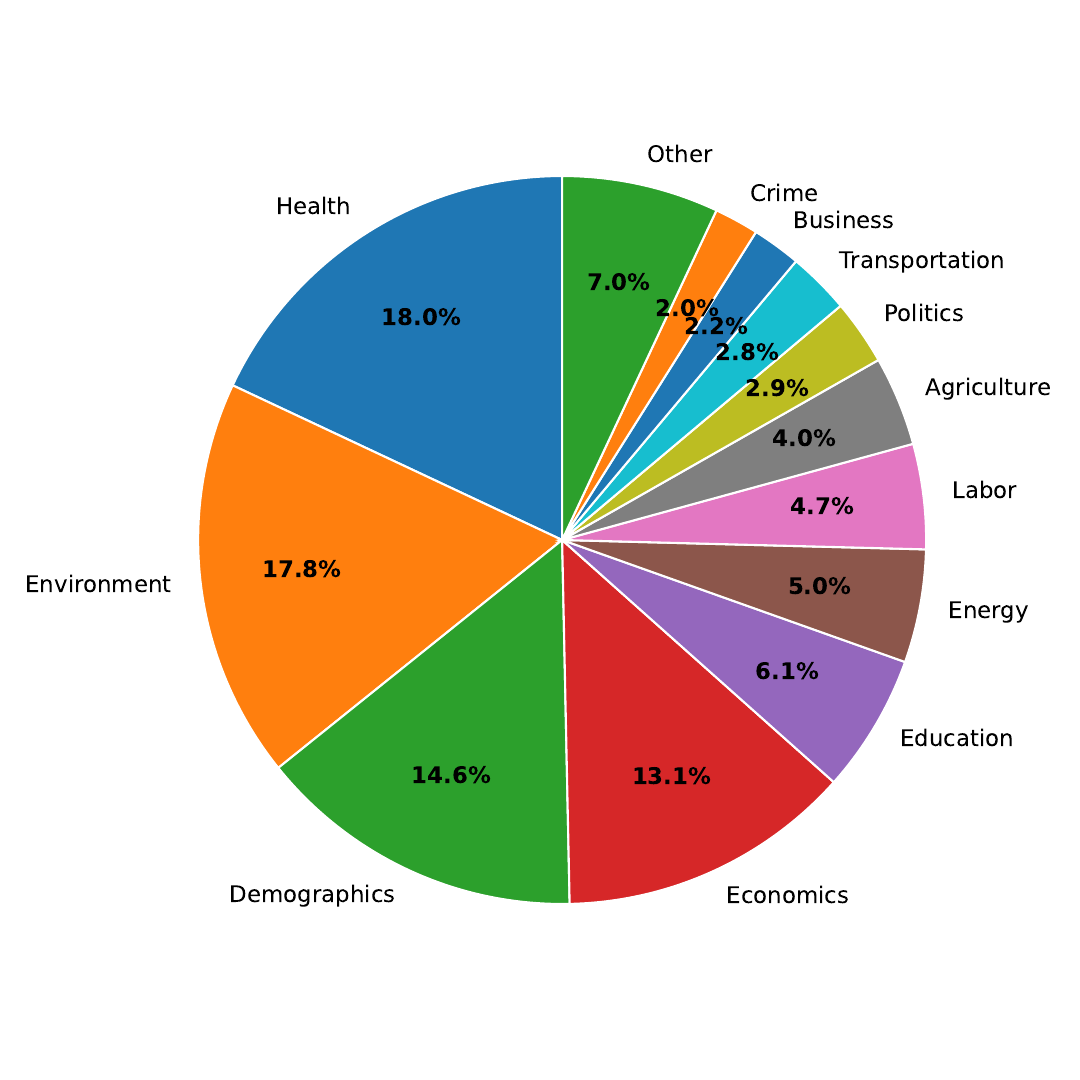}
    \caption{Topic distribution of VIS-GEN. The 12 most frequent domains are shown explicitly; remaining categories are aggregated as Other.}
    \label{fig:topic-statistics}
    \vspace{-2mm}
    % \vskip -2ex
\end{figure}

 \begin{figure}[t!]
    \includegraphics[width=\textwidth]{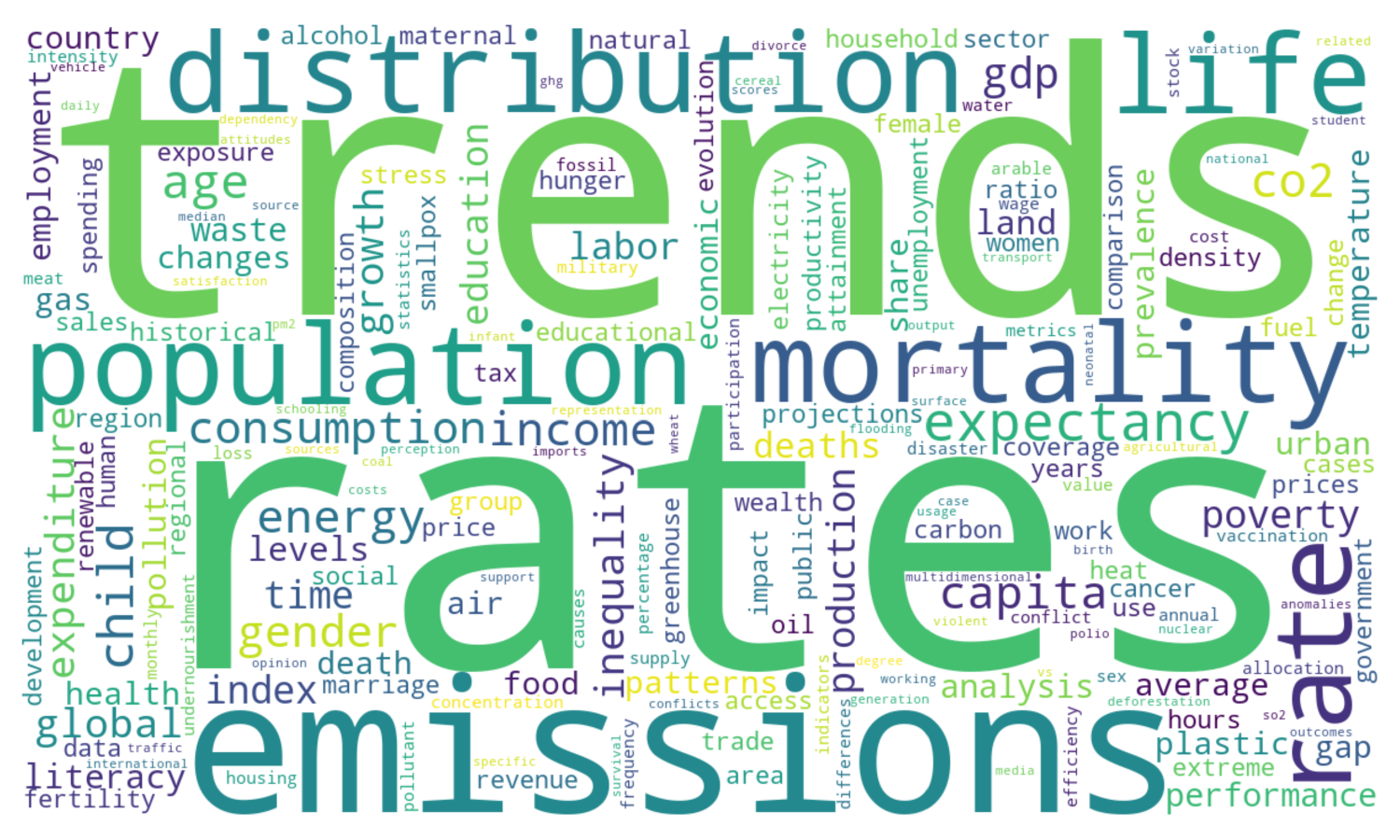}
    \caption{Word cloud illustrating the distribution of fine-grained analytical topics in VIS-GEN. Term size is proportional to frequency of occurrence.}
    \label{fig:topic-statistics1}
    \vspace{-2mm}
    % \vskip -2ex
\end{figure}

\begin{table}[t!]
\centering
\small
\renewcommand{\arraystretch}{1.3} 

\caption{Distribution of LLM-generated candidate queries used for dataset expansion.}
\label{tab:model-distribution}

% Use tabular* to force the table to span the full column width
\begin{tabular*}{\columnwidth}{l @{\extracolsep{\fill}} r}
\toprule
\textbf{Model} & \textbf{Count (\%)} \\
\midrule
GPT-4o            & 900 (33.33\%) \\
\rowcolor{gray!10}
GPT-5.2           & 600 (22.22\%) \\
Claude Sonnet 4.5 & 600 (22.22\%) \\
\rowcolor{gray!10}
Gemini 2.5 Pro    & 600 (22.22\%) \\
\midrule
\textbf{Total}    & \textbf{2,700 (100.00\%)} \\
\bottomrule
\end{tabular*}
\end{table}

\subsection{Methodology}
\label{sec:candidate-selection}
\subsubsection{Candidate Selection Details}

The candidate selection score is computed as:
\begin{align}
\text{Score} =\;& 0.3 \cdot \text{WidgetMatch} 
              + 0.3 \cdot \text{IntentCoverage} \nonumber \\
              &+ 0.2 \cdot \text{VizFit}
              + 0.2 \cdot \text{Feasibility}.
\end{align}

The weights are chosen to prioritize analytical intent and interaction completeness, as deficiencies in \textit{WidgetMatch} and \textit{IntentCoverage} constitute the dominant failure modes observed in preliminary analysis. Visualization fit and feasibility remain essential, but are treated as secondary once the core analytical requirements and interaction structure are satisfied. We further analyze the robustness of these weights through a sensitivity study on 250 stratified samples in Table~\ref{tab:weight-sensitivity}.

\begin{table}[t]
\centering
\small
\begin{tabular}{lccccc}
\toprule
Setting & Widget & Intent & VizFit & Feas. & Final Pass \\
\midrule
Default & 0.30 & 0.30 & 0.20 & 0.20 & 37.6 \\
Widget-heavy & 0.40 & 0.25 & 0.20 & 0.15 &39.6  \\
Intent-heavy & 0.25 & 0.40 & 0.20 & 0.15 &36.4  \\
Viz-heavy & 0.25 & 0.25 & 0.35 & 0.15 & 35.2 \\
Feasibility-heavy & 0.25 & 0.25 & 0.15 & 0.35 & 35.6 \\
Uniform & 0.25 & 0.25 & 0.25 & 0.25 &  37.2\\
\bottomrule
\end{tabular}
\caption{Sensitivity analysis of candidate selection weights in Stage 3, based on 250 stratified samples.}
\label{tab:weight-sensitivity}
\end{table}

\subsubsection{Interaction Reactivity Validation} We initially implemented a Playwright-based programmatic interaction test to verify widget reactivity by modifying control values and checking whether the visualization updated accordingly. While this test works in most cases, it can undercount valid interfaces because generated outputs may use heterogeneous or non-standard widget implementations that are difficult to detect automatically. To complement this, we manually audited all PASS samples from 1,000 multi-stage generations. Among these, 502 samples satisfied the final PASS criterion, and 493/502 (98.21\%) exhibited correct widget--visualization linkage. We also observe that broken linkage typically manifests as either invalid/empty rendered data or visibly broken/inconsistent visualizations, which are already captured by our data validity, readability, visual quality, and correctness checks. Therefore, for samples that already fail the final PASS criterion, additional reactivity testing does not change the overall pass outcome. These results suggest that our final PASS criterion provides a more reliable estimate of functional interactivity than a purely automatic Playwright reactivity test, which may falsely penalize valid interfaces due to missed or non-standard widgets.

\subsubsection{Structured Visualization Interface Representation (SVIR)}
\label{app:svir_schema}

We define the Structured Visualization Interface Representation (SVIR) as a structured JSON representation that captures the semantic requirements needed to generate an interactive visualization interface. Formally, an SVIR instance can be represented as:
\[
\mathrm{SVIR} = \{I, D, V, W, C\},
\]
where \(I\) denotes the analytical intent, \(D\) denotes relevant data-schema information, \(V\) denotes visualization design requirements, \(W\) denotes interaction requirements, and \(C\) denotes interface-level constraints.

\begin{itemize}
    \item \textbf{Analytical intent (\(I\))}: target variables, grouping or comparison dimensions, temporal scope, and required analytical operations.
    \item \textbf{Data schema (\(D\))}: relevant fields from the input table and their semantic or data types.
    \item \textbf{Visualization requirements (\(V\))}: required or preferred visual encodings, such as mark type and field-to-channel mappings.
    \item \textbf{Interaction requirements (\(W\))}: required interaction functions, such as filtering, temporal navigation, comparison, annotation, encoding modification, or editing.
    \item \textbf{Constraints (\(C\))}: validity conditions, including valid field usage, required interactions, schema compatibility, and non-empty executable rendering.
\end{itemize}

SVIR specifies the interface at a semantic level: it captures what the generated interface must support without committing to a particular concrete UI implementation. This allows the subsequent stages to generate multiple candidate interfaces, evaluate them under shared constraints, and compile the selected design into an executable visualization interface.

\paragraph{Illustrative SVIR Example.}
A formal specification and illustrative example of the SVIR schema are provided in Figure~\ref{fig:svir-schema}.

\begin{figure*}[t!]
\centering

\begin{tcolorbox}[
    colback=gray!5!white,
    colframe=gray!60!black,
    title=\textbf{SVIR Schema},
    fonttitle=\bfseries\small,
    boxrule=0.8pt,
    arc=2pt,
    left=3pt, right=3pt, top=3pt, bottom=3pt
]
\small
\begin{verbatim}
{
  "intent": {
    "target_variables": ["value"],
    "grouping_variables": ["country"],
    "temporal_scope": "year",
    "operations": ["compare", "filter", "trend"]
  },
  "data_schema": {
    "fields": [
      {"name": "year", "type": "temporal"},
      {"name": "country", "type": "nominal"},
      {"name": "value", "type": "quantitative"}
    ]
  },
  "visualization": {
    "mark": "line",
    "encodings": {
      "x": "year",
      "y": "value",
      "color": "country"
    }
  },
  "interactions": [
    {"function": "filtering", "target": "country"},
    {"function": "temporal_navigation", "target": "year"}
  ],
  "constraints": {
    "use_valid_fields_only": true,
    "requires_non_empty_rendering": true,
    "required_interactions": [
      "filtering",
      "temporal_navigation"
    ]
  }
}
\end{verbatim}
\end{tcolorbox}

\vspace{-2mm}
\caption{
\textbf{Illustrative SVIR schema.}
SVIR represents the semantic requirements of an interactive visualization interface, including analytical intent, data schema, visualization design, interaction requirements, and feasibility constraints.
}
\label{fig:svir-schema}
\vspace{-4mm}
\end{figure*}

\subsection{Cost Analysis}
\label{app:cost-analysis}

Direct generation requires a single model call per sample. In contrast, the multi-stage framework requires approximately six calls, depending on the number of generated candidates and refinement iterations, representing a bounded overhead. These calls are not uniform in cost: early stages such as intent extraction, SVIR construction, and critique produce short structured outputs, while most tokens are consumed by final HTML/Vega-Lite generation and refinement. Despite this overhead, the framework substantially improves reliability, increasing GPT-4o performance from 12.30\% to 37.38\% and Claude-4.5-Opus from 41.50\% to 57.40\%. This cost is acceptable for interactive UI generation, which is typically a latency-tolerant, high-value task where correctness and usability are more important than marginal token cost. Moreover, structured GPT-4o approaches the performance of direct Claude-4.5-Opus, suggesting that the framework can partly offset cost by enabling a cheaper base model to match a stronger single-pass baseline.

\subsection{Additional Results}
\label{sec:additional-results}
Further breakdowns of model performance by query type, complexity, reasoning depth, and number of widgets are provided in \Cref{tab:model-dimension-breakdown}.

\subsection{Human--Judge Agreement and Evaluator Robustness}
\label{sec:human_judge_agreement}

\paragraph{Correlation with Human Ratings.}
We compare human ratings with LLM-judge scores on 1000 stratified samples and observe strong agreement across all evaluation metrics (Table~\ref{tab:human-judge-corr}), with consistently high Pearson ($r$) and Spearman ($\rho$) correlations.

\paragraph{Stability Across Runs.}
To assess robustness, we rerun the LLM judge four times on the 1,000 multi-stage samples and find that the final pass rate remains within $\pm 1$ percentage point across runs, indicating stable and reproducible evaluator behavior.

\begin{table}[t!]
\centering
\small

\setlength{\tabcolsep}{12pt} 

\renewcommand{\arraystretch}{1.3} 

\caption{Correlation between human ratings and LLM-judge scores on 2000 stratified samples.}
\label{tab:human-judge-corr}

\begin{tabular}{lcc}
\toprule
\textbf{Metric} & \textbf{Pearson ($r$)} & \textbf{Spearman ($\rho$)} \\
\midrule
Executability     & 1.00 & 1.00 \\
\rowcolor{gray!10}
Widget Match      & 0.92 & 0.93 \\
Data Validity     & 0.95 & 0.94 \\
\rowcolor{gray!10}
Readability       & 0.86 & 0.88 \\
Visual Quality    & 0.91 & 0.91 \\
\rowcolor{gray!10}
Chart Correctness & 0.85 & 0.87 \\
\bottomrule
\end{tabular}
\end{table}

\subsection{Human Review and Revision Protocol for LLM-Generated Queries}
\label{app:human_review_protocol}

To reduce artifacts from LLM-assisted dataset expansion, every LLM-generated candidate query underwent a mandatory human review and revision process before inclusion in VIS-GEN. The goal of this process was to ensure that each query was natural, analytically meaningful, grounded in the corresponding table schema, and paired with valid visualization and interaction annotations.

\paragraph{Candidate pool.}
Starting from the 400 expert-authored seed queries, we generated additional candidate queries using four heterogeneous LLMs: GPT-4o, GPT-5.2, Claude Sonnet 4.5, and Gemini 2.5 Pro. To encourage the generated queries to reflect the style, diversity, and analytical intent of the human-authored seed set, each generation prompt included three randomly selected seed examples as few-shot demonstrations. These examples were sampled from the initial human-written query pool and varied across task type, interaction explicitness, visualization type, and widget function.

Query generation followed a multi-step agentic process. First, a generator model produced a candidate query together with its intended visualization and interaction annotations. Then, a separate critique step evaluated whether the candidate satisfied the required quality criteria: schema grounding, analytical validity, interaction correctness, visualization compatibility, naturalness and realism, clarity and unambiguity, and contribution to dataset diversity. If the critique identified issues, the candidate was revised and checked again. This generate--critique--revise process was repeated for multiple rounds, and only the final candidate from this iterative process was passed to human review. This ensured that the initial candidate pool had already been filtered for basic quality before manual annotation.

Each candidate consisted of a natural language query, the associated table, the task setting, the intended visualization type, the required widget/interface annotations, and relevant metadata.

\paragraph{Independent human review.}
Each LLM-generated candidate was independently reviewed by two annotators with experience in data visualization and natural-language-based analytical tasks. Annotators examined the query together with the corresponding table schema, available fields, expected visualization type, widget labels, and interaction annotations. For editing tasks, annotators also reviewed the initial Vega-Lite visualization.

Annotators made a binary decision: \textsc{Accept} if the candidate satisfied all review criteria, and \textsc{Revise/Reject} if one or more criteria failed. The seven review criteria were:

\begin{enumerate}[leftmargin=*]
    \item \textbf{Schema grounding.} The query must refer only to fields, values, or analytical concepts supported by the corresponding table.
    \item \textbf{Analytical validity.} The query must express a meaningful visualization-oriented analytical goal, such as filtering, comparison, temporal exploration, ranking, annotation, aggregation, or visualization editing.
    \item \textbf{Interaction correctness.} The annotated widgets must correctly support the analytical intent of the query.
    \item \textbf{Visualization compatibility.} The expected chart type and interaction design must be feasible for the table schema and data types.
    \item \textbf{Naturalness and realism.} The query should resemble a realistic user request for data exploration or visualization editing, rather than a synthetic or template-like instruction.
    \item \textbf{Clarity and unambiguity.} The intended analytical operation must be clear enough for a model or human to determine the required visualization and interactions.
    \item \textbf{Difficulty and diversity.} The candidate should contribute useful coverage across task type, interaction explicitness, reasoning depth, visualization type, and widget function, without being a near-duplicate of existing examples.
\end{enumerate}

A candidate was accepted only when all seven criteria were satisfied. If any criterion failed, the candidate was marked for revision or rejection.

\paragraph{Revision and disagreement resolution.}
The initial direct-accept agreement between the two annotators was 82.7\%. A case was marked for further review whenever either annotator indicated that the candidate should be revised or rejected rather than directly accepted. This resulted in 467 cases requiring additional review.
% The initial direct-accept agreement between the two annotators was 81.7\%. A case was marked for further review whenever either annotator indicated that the candidate should be revised or rejected rather than directly accepted. This resulted in 494 cases requiring additional review.

All 467 cases requiring additional review were manually revisited by both annotators. During this stage, the annotators reviewed the query, table schema, visualization specification, and widget annotations. Each case was revised, edited, or updated to ensure that it satisfied the seven review criteria: schema grounding, analytical validity, interaction correctness, visualization compatibility, naturalness and realism, clarity and unambiguity, and difficulty/diversity. Revisions included clarifying ambiguous wording, correcting unsupported field references, updating widget annotations, adjusting visualization types, or rewriting the query when needed.

After this revision stage, the updated candidates were cross-checked again by both annotators and assigned a final pass/fail decision. In the second review pass, a candidate was removed if either annotator determined that it should still be rejected. This resulted in 58 discarded candidates. These cases were removed to maintain a conservative quality standard.

\paragraph{Final inclusion rule.}
A candidate was included in VIS-GEN only if it satisfied all of the following conditions:

\begin{enumerate}[leftmargin=*]
    \item The query was natural and analytically meaningful.
    \item All referenced fields and values were grounded in the table.
    \item The expected visualization was compatible with the task and data types.
    \item The required widgets correctly supported the intended interaction.
    \item The task labels and metadata were internally consistent.
    \item The candidate was accepted by both annotators, either during the initial independent review or after disagreement resolution.
\end{enumerate}

This conservative protocol ensures that VIS-GEN is not simply a collection of LLM-generated queries, but a human-validated benchmark for evaluating whether models can infer and implement interactive visualization interfaces from realistic analytical language.

\subsection{Visualization Interface Generation Comparison}
\label{sec:vis_interface_comparison}

\begin{figure*}[t!]
    \centering
    \includegraphics[width=.98\textwidth]{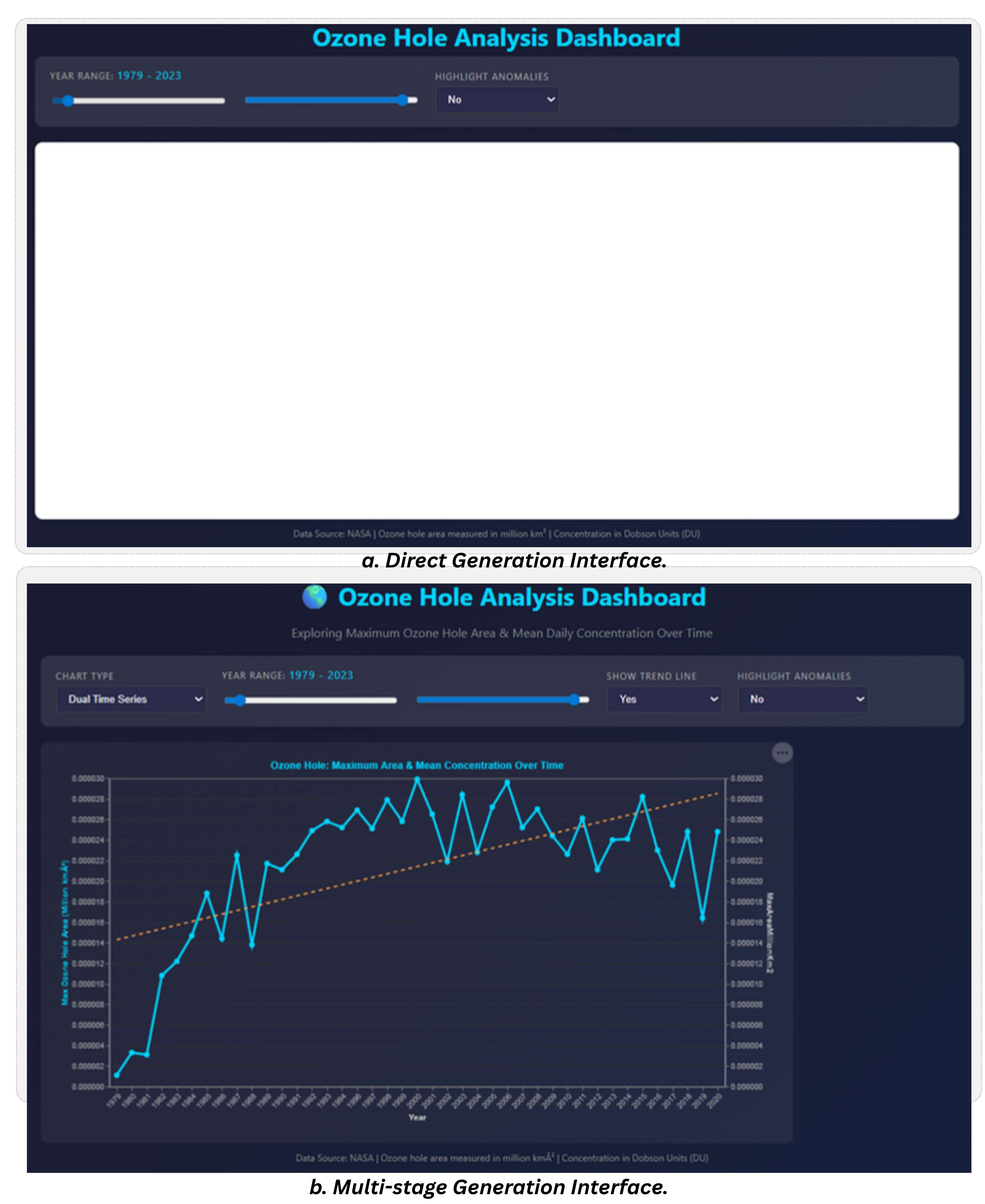}
\caption{
\textbf{Visualization Interface Generation Comparison.}
\textbf{(a) Direct Generation Interface} produced in a single-pass setting, in which the chart fails to render due to a data validation failure, despite the presence of a complete interface layout. 
\textbf{(b) Multi-stage Generation Interface} produced by the proposed framework, which iteratively refines candidate designs through structured validation, critique, and selection, yielding improved interaction completeness, execution reliability, and visual–semantic alignment.
\vspace{-5mm}
}
    \label{fig:vis-comparison}
    \vspace{-3mm}
\end{figure*}

We provide a qualitative comparison between direct and multi-stage interface generation to illustrate the practical impact of the proposed framework.
Figure~\ref{fig:vis-comparison} shows a representative example in which \textbf{Claude-4.5-Opus} under direct generation produces a visually complete interface that fails to render due to a data validation error, while the proposed multi-stage framework successfully recovers through structured critique, validation, and refinement, yielding a functional and semantically aligned interactive visualization.

\subsection{Error analysis}
\label{sec:error_analysis}

\begin{figure*}[t!]
    \centering
    \includegraphics[width=.98\textwidth]{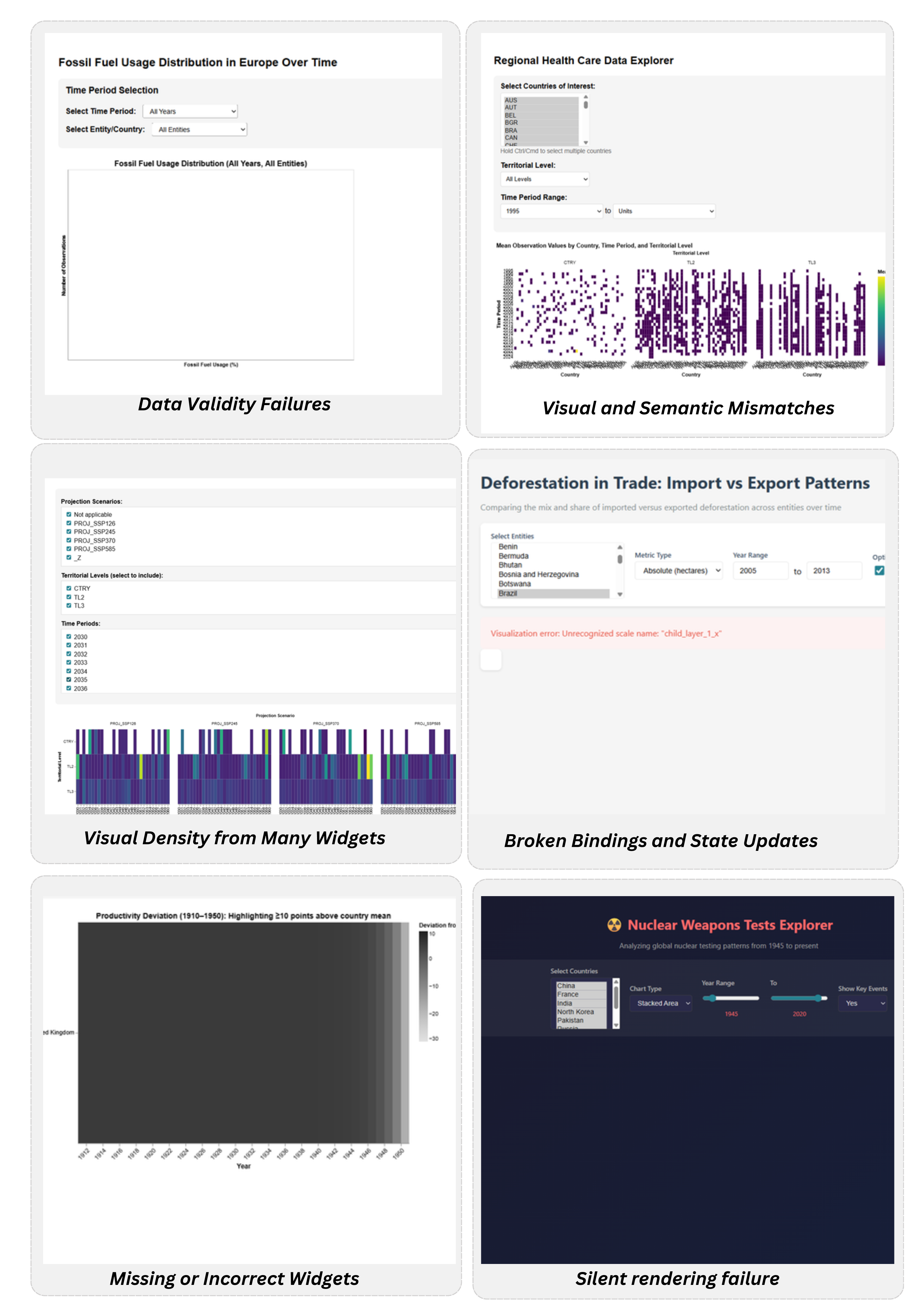}
\caption{Representative failure modes in interactive visualization generation, including data validity failures, visual and semantic mismatches, visual density from excessive widgets, broken bindings and state updates, missing or incorrect widgets, and silent rendering failures.
\vspace{-5mm}
}    

    \vspace{-3mm}
    \label{fig:error-taxonomy}
    % \vskip -2ex
\end{figure*}

We conduct a qualitative error analysis to characterize the dominant failure modes in language-driven interactive visualization generation.
Figure~\ref{fig:error-taxonomy} summarizes representative error patterns observed across models, including data validity failures, visual and semantic mismatches, excessive interface complexity from dense widgets, broken bindings and state updates, missing or incorrect widgets, and silent rendering failures.
These observations complement the quantitative results and highlight the key challenges that remain for reliable interface-level reasoning.

\subsection{Prompt Design}
\label{sec:prompt-design}

We include representative sample prompts for the proposed multi-stage interactive generation framework in this section, and will release the full set of prompts, evaluation scripts, and implementation code as part of the public repository upon publication.

\begin{figure*}[t]
    \centering
    
    % --- STANDARD PROMPT BOX ---
    \begin{tcolorbox}[
        colback=gray!5!white,
        colframe=gray!60!black,
        title=\textbf{Stage 1 (a): Intent Extraction Prompt},
        fonttitle=\bfseries\small,
        boxrule=0.8pt,
        arc=2pt,
        left=2pt, right=2pt, top=2pt, bottom=2pt
    ]
    \fontfamily{cmtt}\selectfont\small
    You are designing an interactive analytical visualization system. \\
    
    Given: \\
    (1) a dataset schema \\
    (2) a natural language user query \\
    
    Your job is to extract the analytical goals and the interactive controls required to support the analysis. \\
    
    Think in terms of interaction needs, not UI implementation. \\

    Return only valid JSON in the following format: \\

    \begin{verbatim}
{
  "objectives": [
    {
      "task": "<short analytical goal>",
      "priority": 1-5,
      "rationale": "<why this goal is necessary>"
    }
  ],

  "visual_analytic_form": {
    "family": "comparison|trend|distribution|
    relationship|composition|geospatial|
    ranking|anomaly|..."

  },

  "interaction_requirements": [
    {
      "operation": "filter|range|threshold|time_navigation|
      compare|sort|aggregate|search|
      focus|highlight|...."
,
      "field": "<dataset column>",
      "importance": 1-5,
      "rationale": "<why this control is required>"
    }
  ],

  "time_field": "<dataset column|null>"
}
    \end{verbatim}

    Rules: \\
    -- Use ONLY dataset column names for fields. \\
    -- Do NOT mention widgets, UI, HTML, or implementation in the JSON. \\
    -- Include only controls that materially help answer the query. \\
    -- If the query implies time, set \texttt{time\_field}. \\

    \end{tcolorbox}
    
    \vspace{0.2cm}

    \caption{
    \textbf{Stage 1(a) Prompt: Intent Extraction.}
    The system prompt used in the first stage of the proposed multi-stage framework, which extracts high-level analytical objectives and required interaction operations from a dataset schema and natural language query.
    }
    \label{fig:prompt-intent}
\end{figure*}

\begin{figure*}[t]
    \centering
    
    % --- STANDARD PROMPT BOX ---
    \begin{tcolorbox}[
        colback=gray!5!white,
        colframe=gray!60!black,
        title=\textbf{Stage 1(b): SVIR Construction Prompt},
        fonttitle=\bfseries\small,
        boxrule=0.8pt,
        arc=2pt,
        left=2pt, right=2pt, top=2pt, bottom=2pt
    ]
    \fontfamily{cmtt}\selectfont\small
    Convert the inferred intents into SVIR (Structured Visualization Interface Representation) that separates semantics from implementation. \\

    If an EXISTING\_VEGA\_LITE\_SPEC block was provided before this prompt: \\
    -- Treat it as the current implementation that the user wants to edit or refine. \\
    -- Align visualization\_intent and constraints with that spec when reasonable. \\

    Return only valid JSON in the following format: \\

    \begin{verbatim}
{
  "visualization_intent": {
    "task": "<one sentence summary of what the visualization must show>",
    "visual_family": "<must exactly match Stage1.visual_analytic_form.family>",
    "preferred_marks": ["point|bar|line|area|rect|text|circle|square|rule"|.."],
    "required_encodings": {
      "x": "<field|null>",
      "y": "<field|null>",
      "color": "<field|null>",
      "size": "<field|null>",
      "shape": "<field|null>",
      "facet": "<field|null>"
    },
    "encoding_rationale": "<why these specific encodings best serve the task>"
  },

  "exploration_intents": [
    {
      "type": "range_filter|category_filter|boolean_filter|threshold_filter|
      time_navigation|search|sort|aggregate|compare|focus|
      drilldown|rollup|highlight|navigate|...",
      "field": "<dataset column>",
      "importance": 1-5,
      "operation_semantics": "<what analytical question does this enable>",
      "expected_behavior": "<what should happen when the user interacts>"
    }
  ],

  "constraints": {
    "field_types": {
      "<field>": "quantitative|temporal|nominal|ordinal|text|
      boolean|id|geojson"
    },
    "encoding_constraints": {
      "x_type": "quantitative|temporal|nominal|ordinal|null",
      "y_type": "quantitative|temporal|nominal|ordinal|null",
      "color_type": "quantitative|nominal|ordinal|null"
    }
  }
}
    \end{verbatim}

    Rules: \\
    -- Stay semantic; do not use UI or widget terminology. \\
    -- Use only dataset column names for fields. \\
    -- Use temporal fields only for time\_navigation. \\
    -- Ensure required\_encodings respect field\_types. \\

    \end{tcolorbox}
    
    \vspace{0.2cm}

    \caption{
    \textbf{Stage 1(b) Prompt: SVIR Construction.}
    The system prompt used to translate extracted analytical intent into a structured semantic specification that constrains visualization design and interaction behavior before interface synthesis.
    }
    \label{fig:prompt-svir}
\end{figure*}

\begin{figure*}[t]
    \centering
    
    % --- STANDARD PROMPT BOX ---
    \begin{tcolorbox}[
        colback=gray!5!white,
        colframe=gray!60!black,
        title=\textbf{Stage 2 Prompt: Candidate Interface Generation},
        fonttitle=\bfseries\small,
        boxrule=0.8pt,
        arc=2pt,
        left=2pt, right=2pt, top=2pt, bottom=2pt
    ]
    \fontfamily{cmtt}\selectfont\small
    You are designing alternative interactive visualization interfaces. \\

    Your job is to propose multiple interface designs that faithfully implement the
    semantic requirements already defined in SVIR. \\
    Do NOT invent new requirements. \\

    Return only valid JSON in the following format: \\

    \begin{verbatim}
{
  "candidates": [
    {
      "id": "C1",
      "chart": {
        "mark": "<one of: point|bar|line|area|rect|tick|circle|square|rule|..>",
        "encodings": {
          "x": "<field|null>",
          "y": "<field|null>",
          "color": "<field|null>",
          "facet": "<field|null>"
        },
        "description": "<short explanation of the design>"
      },
      "controls": [
        {
          "control_type": "range|time|category|boolean|threshold|
          search|sort|aggregate|compare|focus|
          drilldown|rollup|highlight|navigate}...",
          "field": "<dataset column>",
          "importance": 1-5,
          "notes": "<short>"
        }
      ]
    }
  ]
}
    \end{verbatim}

    Requirements: \\
    -- Generate exactly four candidates. \\
    -- Each candidate must implement all required\_encodings from SVIR.visualization\_intent. \\
    -- Each candidate must implement all SVIR.exploration\_intents. \\

    Controls: \\
    -- Use only the interactions specified in SVIR.exploration\_intents and Stage-1. \\
    -- Do not introduce new interaction types. \\
    -- Map each exploration\_intent to exactly one control. \\
    -- For any exploration\_intent with importance at least 4, the control must appear in every candidate. \\

    Diversity: \\
    -- All four candidates must be structurally different. \\
    -- No two candidates may share the same chart and control configuration. \\

    Safety and validity: \\
    -- Use only dataset columns for fields. \\
    -- Ensure encodings respect field types from SVIR.constraints. \\
    -- Do not produce redundant controls. \\
    -- Do not include UI or widget names; controls are abstract. \\

    The goal is to explore multiple high-quality interface designs while strictly
    respecting the intent and interaction requirements. \\

    \end{tcolorbox}
    
    \vspace{0.2cm}

    \caption{
    \textbf{Stage 2 Prompt: Candidate Interface Generation.}
    The system prompt used to generate multiple structurally diverse candidate interfaces that satisfy the semantic intent and interaction constraints defined in the previous stages.
    }
    \label{fig:prompt-candidates}
\end{figure*}

\begin{figure*}[t]
    \centering
    
    % --- STANDARD PROMPT BOX ---
    \begin{tcolorbox}[
        colback=gray!5!white,
        colframe=gray!60!black,
        title=\textbf{Stage 3 Prompt: Candidate Critique and Evaluation},
        fonttitle=\bfseries\small,
        boxrule=0.8pt,
        arc=2pt,
        left=2pt, right=2pt, top=2pt, bottom=2pt
    ]
    \fontfamily{cmtt}\selectfont\small
    Evaluate each candidate using the following criteria: \\

    1) widget\_match: 1 if the candidate includes all interaction\_requirements from Stage-1 and all exploration\_intents from SVIR, else 0 \\
    2) intent\_coverage (1–5): weighted by objectives priority \\
    3) viz\_fit (1–5): how well chart structure matches visualization\_intent.visual\_family from SVIR \\
    4) feasibility (1–5): field type compatibility, schema validity, encoding correctness \\

    Return only valid JSON in the following format: \\

    \begin{verbatim}
{
  "evaluations": [
    {
      "id": "C1",
      "widget_match": 0 or 1,
      "intent_coverage": 1-5,
      "viz_fit": 1-5,
      "feasibility": 1-5,
      "issues": ["<short>", "..."]
    }
  ]
}
    \end{verbatim}

    Rules: \\
    -- Reject candidates with missing required encodings or invalid field types. \\

    \end{tcolorbox}
    
    \vspace{0.2cm}

    \caption{
    \textbf{Stage 3 Prompt: Candidate Critique and Evaluation.}
    The system prompt used to assess candidate interfaces with respect to intent fulfillment, interaction coverage, visualization fit, and execution feasibility under strict rejection constraints.
    }
    \label{fig:prompt-critique}
\end{figure*}

\begin{figure*}[t]
    \centering
    
    % --- STANDARD PROMPT BOX ---
    \begin{tcolorbox}[
        colback=gray!5!white,
        colframe=gray!60!black,
        title=\textbf{Stage 4 Prompt: HTML Generation},
        fonttitle=\bfseries\small,
        boxrule=0.8pt,
        arc=2pt,
        left=2pt, right=2pt, top=2pt, bottom=2pt
    ]
    \fontfamily{cmtt}\selectfont\small
    You are a data visualization system. \\

    Your task is to generate a complete, self-contained HTML file that implements the FINAL\_PLAN
    as a correct and interactive Vega-Lite visualization. \\

    Core constraints: \\
    -- Use correct data types: temporal only for true dates, quantitative for numeric, nominal or ordinal for categorical. \\
    -- Implement the FINAL\_PLAN exactly: chart encodings, mark type, and all specified controls. \\
    -- The dataset must be loaded using the provided dataset path only. \\
    -- The system must be robust under interaction and handle rendering errors. \\
    -- Output only the final HTML document with no explanations. \\
    ... \\

    Reference structure: \\
    Follow the general structure of the provided example HTML. \\
    ... \\

    Task context: \\
    TASK TYPE: \{QUERY\_TYPE\} \\
    USER QUERY: \{QUERY\} \\
    DATASET FILE: \{DATASET\_PATH\} \\
    COLUMNS: \{COLUMN\_LIST\} \\
    FINAL\_PLAN (JSON): \{FINAL\_PLAN\} \\

    \end{tcolorbox}
    
    \vspace{0.2cm}

    \caption{
    \textbf{Stage 4 Prompt: HTML Generation.}
    The system prompt used to synthesize a complete interactive HTML visualization that faithfully implements the final structured design plan.
    }
    \label{fig:prompt-html}
\end{figure*}

\begin{figure*}[t]
    \centering
    
    % --- STANDARD RUBRIC BOX ---
    \begin{tcolorbox}[
        colback=gray!5!white,
        colframe=gray!60!black,
        title=\textbf{Evaluation Scoring Rubric},
        fonttitle=\bfseries\small,
        boxrule=0.8pt,
        arc=2pt,
        left=2pt, right=2pt, top=2pt, bottom=2pt
    ]
    \fontfamily{cmtt}\selectfont\small
    
    \textbf{Task 1 — Widget Match (Binary)} \\
    Decide whether the required widgets or reasonable functional equivalents appear in the interface and meaningfully support the analytical goal stated in the query. \\
    Score: \\
    -- 1: All required widgets (or clearly equivalent interactive mechanisms) are present and correctly support the task. \\
    -- 0: One or more required widgets are missing, ineffective, or unrelated to the analytical goal. \\

    \textbf{Task 2 — Data Validity (Binary)} \\
    Decide whether the visualization displays valid plotted data. \\
    Score: \\
    -- 1: The chart shows actual data marks (e.g., bars, points, lines) and responds correctly to interactions. \\
    -- 0: The chart is empty, broken, shows only axes, or displays error placeholders. \\

    \textbf{Task 3 — Readability (1–5)} \\
    Evaluate how easily a human can understand and use the interface. \\
    Score: \\
    -- 1: Unreadable; cluttered, confusing, mislabeled, or unusable. \\
    -- 2: Major readability problems; difficult to follow with frequent confusion. \\
    -- 3: Acceptable; understandable with some effort. \\
    -- 4: Clear and well-organized; easy to follow. \\
    -- 5: Exceptionally clear; intuitive layout and labeling. \\

    \textbf{Task 4 — Visual Quality (1–5)} \\
    Evaluate the aesthetic quality and professional presentation of the interface. \\
    Score: \\
    -- 1: Very poor; messy, misaligned, or visually broken. \\
    -- 2: Weak design; noticeable visual issues. \\
    -- 3: Adequate; visually acceptable. \\
    -- 4: High quality; clean and polished. \\
    -- 5: Excellent; publication-ready professional design. \\

    \textbf{Task 5 — Correctness (1–5)} \\
    Evaluate whether the visualization and interactions correctly address the user’s query. \\
    Score: \\
    -- 1: Does not answer the query at all. \\
    -- 2: Barely related; mostly incorrect or misleading. \\
    -- 3: Partially answers the query. \\
    -- 4: Mostly correct and useful. \\
    -- 5: Fully answers the query accurately and robustly. \\

    \end{tcolorbox}
    
    \vspace{0.2cm}

    \caption{
    \textbf{Evaluation Rubric.}
    Scoring criteria  to assess interaction completeness, data validity, readability, visual quality, and analytical correctness of generated visualization interfaces.
    }
    \label{fig:scoring-rubric}
\end{figure*}

\end{document}